\documentclass{beingbeyond}

\usepackage{amsmath}
\usepackage{amssymb}
\usepackage{booktabs}
\usepackage{xspace}
\usepackage{subcaption}
\usepackage[shortlabels,inline]{enumitem}

\usepackage{placeins}
\usepackage{multirow}
\usepackage{makecell}

\usepackage[capitalize]{cleveref}
\crefname{section}{Sec.}{Secs.}
\Crefname{section}{Section}{Sections}
\crefname{table}{Tab.}{Tabs.}
\Crefname{table}{Table}{Tables}

\makeatletter
\DeclareRobustCommand\onedot{\futurelet\@let@token\@onedot}
\def\@onedot{\ifx\@let@token.\else.\null\fi\xspace}
\def\eg{\emph{e.g}\onedot}

\makeatother

\DeclareMathOperator*{\argmax}{arg\,max}
\DeclareMathOperator*{\argmin}{arg\,min}

\title{S3AL: Scaling 3D Hand-Object Annotation for Mesh-Free Learning}

\author{\\[1.5mm]
\textbf{Weibo Li}$^{1}$ \quad
\textbf{Yuchen Deng}$^{1}$ \quad
\textbf{Sipeng Zheng}$^{2}$ \\
\textbf{Ziheng Xi}$^{2,3}$ \quad
\textbf{Wanpeng Zhang}$^{2}$ \quad
\textbf{Zongqing Lu}$^{1,2,\dagger}$ \\[3mm]
{\fontsize{11}{13}\selectfont $^{1}$Peking University \quad $^{2}$BeingBeyond \quad $^{3}$Tsinghua University}\\[3mm]
{\fontsize{11}{13}\selectfont\url{https://research.beingbeyond.com/s3al}}
}

\begin{document}
\crefname{appendix}{App.}{App.}\Crefname{appendix}{Appendix}{Appendices}
\maketitle
{\begin{NoHyper}\let\thefootnote\relax\footnotetext{$^\dagger$Corresponding author.}\end{NoHyper}}%
\begin{abstract}
Learning from human experience is central to embodied foundation models, yet requires accurate 3D hand-object interaction (HOI) annotations.
Existing methods struggle with in-the-wild videos due to inaccurate reconstruction, costly optimization, or expensive agentic pipelines. 
We present \textbf{S3AL}, a framework aiming to \textbf{S}cale \textbf{3}D hand-object \textbf{A}nnotation for mesh-free \textbf{L}earning, which recovers dense 3D labels from monocular videos at over 10$\times$ lower annotation cost than prior methods. 
S3AL integrates pretrained hand, depth, and mesh priors with VLM-based verification for metric-scale estimation and pose selection, followed by contact-aware optimization, VLM filtering, and temporal refinement.
It outperforms depth-assisted pose trackers and monocular HOI methods across all evaluated metrics, resulting in significantly higher quality while substantially reducing the cost. 
Using S3AL, we annotate over 1M in-the-wild frames to train a feed-forward predictor of mesh-free, object-centric 3D representations. 
Given an RGB image, object mask, and estimated hand poses, the model predicts a canonical object frame with metric pose and extent, refined hand poses, and signed hand-to-object surface distances.
The predictor outperforms depth- and mesh-based baselines and generalizes zero-shot to unseen objects and third-person views.
These results demonstrate that feed-forward visual models can infer structured 3D hand-object geometry without explicit mesh reconstruction, offering a scalable foundation for 3D interaction understanding.

\end{abstract}
\begin{figure}[t]
\centering
\includegraphics[width=0.96\textwidth]{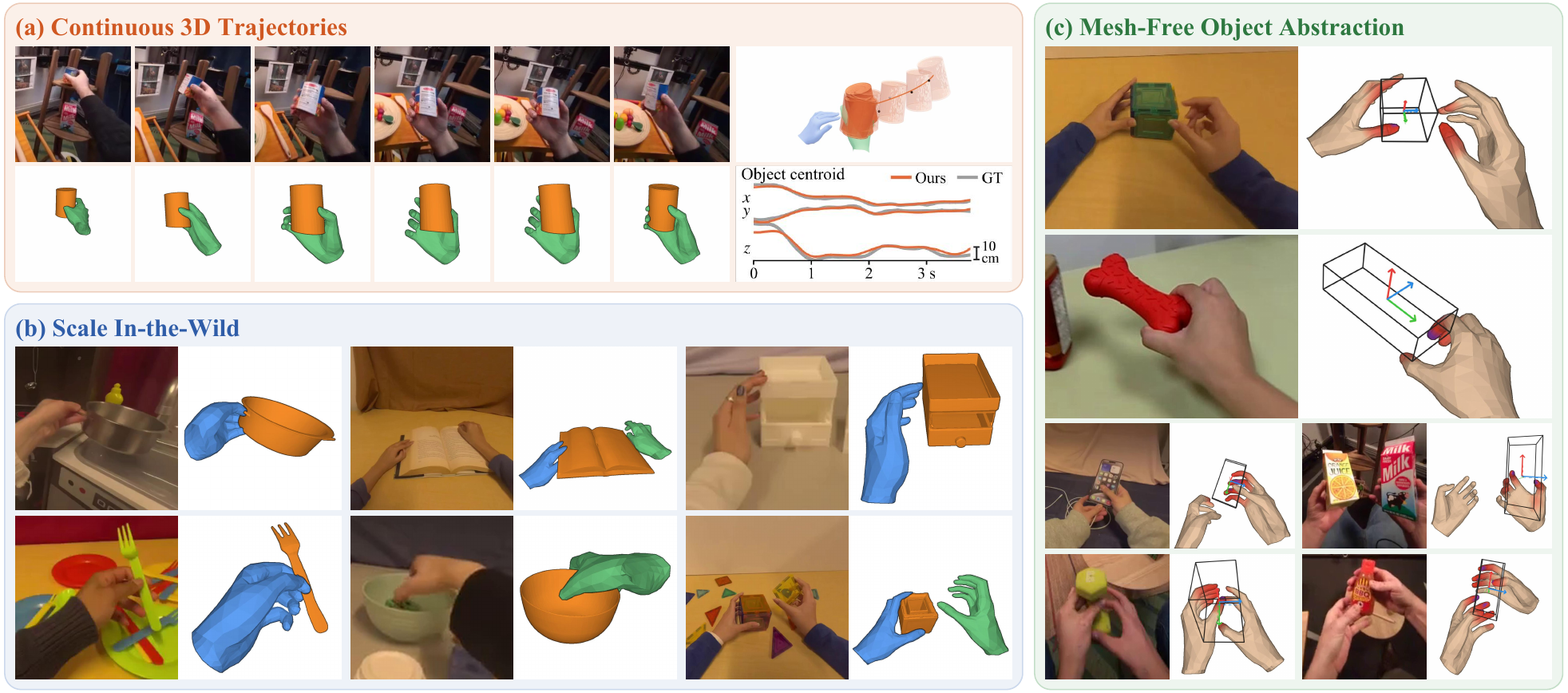}
\caption{
\textbf{(a)} By decoupling VLM verification from contact-aware optimization and temporal smoothing, S3AL recovers dense 3D hand-object labels from monocular video at substantially lower cost than optimization methods. 
\textbf{(b)} We deploy S3AL on in-the-wild videos to curate over 1M reviewed 3D HOI frames. 
\textbf{(c)} Inspired by mesh-free object abstraction, we distill these labels into a feed-forward predictor, which infers a metric canonical frame and box, refined hands, and hand-to-object signed distances (red near contact) without mesh reconstruction.}
\label{fig:teaser}
\end{figure}
\section{Introduction}
\label{sec:intro}

Learning dexterous manipulation from human video~\cite{qin2021dexmv,pmlr-v205-shaw23a,singh2024handobjectinteractionpretrainingvideos,beingbeyond2025beingh0} requires dense 3D supervision of both hands and manipulated objects, including metric object pose, scale, and contact with hands.
Existing datasets provide accurate annotations but are largely confined to controlled laboratory settings, limiting their diversity and requiring calibrated multi-camera rigs and pre-scanned objects~\cite{hampali2020honnotate,chao:cvpr2021,Zhan_2024_CVPR,banerjee2025hot3d}.
In contrast, in-the-wild egocentric videos~\cite{egodex,9879279,Damen2018EPICKITCHENS} are abundant, yet existing hand estimators~\cite{pavlakos2024reconstructing,potamias2024wilor,zhang2025hawor} and large-scale collections such as EgoDex~\cite{egodex} provide only 3D hand trajectories, leaving object geometry and motion largely unlabeled.

Some feed-forward predictors~\cite{chen2026forehoifeedforward3dobject} offer efficient inference but struggle to generalize beyond their training domains.
Optimization-based approaches~\cite{fan2024hold,wang2025magichoi,liu2025easyhoi,paliwal2026doasido} rely on costly per-sequence fitting, often using 2D silhouette constraints and heuristic 3D proximity losses, making them sensitive to depth-initialization drift and unable to filter out reconstruction failures.
More recently, agentic trackers~\cite{mazur2026agenticstar} incorporate vision-language models (VLMs) into iterative optimization to resolve ambiguous 3D hypotheses.
However, repeated VLM queries and expensive optimization remain prohibitive for million-frame annotation.

In this paper, we present \textbf{S3AL} (\cref{fig:teaser}), a scalable data forge to \textbf{S}cale \textbf{3}D hand-object \textbf{A}nnotation for mesh-free \textbf{L}earning. 
S3AL decouples discrete semantic selection from continuous contact-aware refinement. 
Specifically, we exploit pretrained models to provide object masks~\cite{carion2025sam3segmentconcepts}, 3D hand poses~\cite{zhang2025hawor}, hand-calibrated metric depth~\cite{depthanything3}, a per-trajectory canonical mesh~\cite{Chen_2026_CVPR}, and rotation candidates for each frame.
VLM verification then estimates metric scale and selects plausible pose hypotheses at bounded cost.
To resolve the remaining in-grasp pose and depth ambiguity, we leverage dense hand-contact priors~\cite{xi2026tactohandlearninggeometrybaseddense} inferred directly from grasp appearance to refine candidate poses, followed by filtering and smoothing.

Beyond scaling annotation to over 1M in-the-wild frames, S3AL enables mesh-free, object-centric 3D representation learning.
This raises a key question: what 3D target should an embodied model predict? 
Decoding full object meshes from a single occluded view is ill-posed and unnecessarily detailed for manipulation.
Conventional 6D pose estimation requires known CAD models or predefined object categories~\cite{Wang_2019_CVPR}, while vision-based contact predictors remain hand-centric~\cite{jung2025haco,xi2026tactohandlearninggeometrybaseddense}. 
Instead, we introduce a \emph{mesh-free, object-centric abstraction} comprising a canonical object frame with metric pose and extent, refined hand poses in that frame, and signed distances from hand vertices and joints to the object surface. 
Derived directly from geometry with symmetry-aware supervision, this fixed-size representation applies across diverse object shapes and can be predicted within a forward pass given an RGB image, an object mask, and estimated hand poses.
Our contributions in this paper can be summarized as:
\begin{itemize}
    \item We propose S3AL, a scalable monocular 3D HOI data annotation forge that combines geometric priors, contact-aware optimization, and VLM verification. S3AL outperforms existing baselines at over $10\times$ lower cost than optimization-based pipelines.
    \item We curate over 1M reviewed in-the-wild 3D HOI frames from a large-scale egocentric corpus~\cite{egodex}, with only 0.2\% audited gross errors, and will publicly release the annotations to support large-scale embodied learning.
    \item We introduce a mesh-free, object-centric 3D representation for canonical object abstraction. To achieve this, we train a feed-forward predictor that outperforms depth- and mesh-based baselines and generalizes zero-shot to unseen objects and viewpoints.
\end{itemize}

\section{Related Work}
\label{sec:related}

\noindent\textbf{3D Hand-Object Datasets.}
Precise 3D HOI annotations typically come from synthetic grasps~\cite{hasson19_obman}, calibrated multi-view RGB-D rigs with known objects~\cite{hampali2020honnotate,9880464,chao:cvpr2021,Kwon_2021_ICCV,wang2025hocap}, motion-capture or dense camera studios~\cite{fan2023arctic,YangCVPR2022OakInk,Zhan_2024_CVPR,liu2024taco,2024graspnet,fu2025gigahands}, and wearable multi-sensor rigs~\cite{Liu_2022_CVPR,banerjee2025hot3d,rim2026show3d}. 
While accurate, all of these sources rely on synthetic rendering, pre-scanned objects, or specialized capture hardware. 
Conversely, in-the-wild egocentric and Internet collections~\cite{9879279,Damen2018EPICKITCHENS,egodex,9157473} offer broad visual diversity without 3D object geometry, and manual 3D fitting in the wild scales only to small image sets~\cite{Cao2021}. 
Recent data engines~\cite{zhang2025robowheel,egoinfinity_2026} also recover 3D hand-object labels from video, but either target controlled captures~\cite{zhang2025robowheel} or rely on heuristic filters without ground-truth 3D object evaluation~\cite{egoinfinity_2026}. 
In contrast, S3AL scales to over 1M in-the-wild frames with VLM-verified reconstructions and benchmarked 3D accuracy.

\noindent\textbf{Monocular Hand-Object Reconstruction.}
Off-the-shelf 3D hand estimators~\cite{MANO:SIGGRAPHASIA:2017,pavlakos2024reconstructing,potamias2024wilor,zhang2025hawor} generalize well in the wild but ignore manipulated objects. 
Single-image HOI methods regress mesh templates~\cite{hasson20_handobjectconsist}, implicit fields~\cite{ye2022hand,chen2023gsdf,qi2024hoisdf}, or point clouds~\cite{chen2025hort} in hand-relative coordinates; feed-forward video models~\cite{chen2026forehoifeedforward3dobject,li20264dhofhandobjectflowmatching} run fast but degrade out of distribution in the wild. 
Per-sequence optimization methods fit implicit surfaces or meshes over video clips~\cite{9665955,ye2023vhoi,fan2024hold,wang2025magichoi} at high per-clip compute cost, and foundation-model pipelines~\cite{liu2025easyhoi,paliwal2026doasido} optimize noisy cues without automated semantic verification of reconstructed grasps. 
Recent agentic frameworks either pair iterative VLM-guided asset generation with thorough per-clip physics optimization~\cite{shi2026agile} or place a VLM directly inside a render-and-compare pose loop~\cite{mazur2026agenticstar}, at high latency and API cost. 
In contrast, S3AL invokes the VLM at four fixed stages to select and verify rendered hypotheses at bounded cost.

\noindent\textbf{Object Abstractions and Contact Priors.}
Model-based 6D pose trackers~\cite{labbe2022megapose,foundationposewen2024} require a CAD mesh at test time, while model-free RGB-D variants~\cite{lee2025any6d} still reconstruct an explicit proxy mesh from a reference view. 
Category-level estimators~\cite{Wang_2019_CVPR} learn per-category canonical spaces with symmetry-aware losses~\cite{Hodan_2018_ECCV,8885689}, monocular 3D detectors~\cite{9578264,10204028} regress oriented boxes for closed categories, and multi-primitive abstractions~\cite{abstractionTulsiani17,Paschalidou2019CVPR} fit variable-count cuboids or superquadrics for detailed shape rather than fixed-size prediction targets. 
For interaction, prior methods derive contact from 3D surface intersection~\cite{chen2023gsdf,qi2024hoisdf}, heuristic proximity in joint optimization~\cite{9665955,liu2025easyhoi}, or perturbed grasps on known CAD models~\cite{9578455}. 
Recent estimators infer dense hand-surface contact~\cite{jung2025haco} or video interaction fields~\cite{xi2026tactohandlearninggeometrybaseddense} directly from visual appearance. 
However, these predictors remain hand-centric, estimating fields on the MANO surface or refining poses with known meshes without recovering open-world object pose, metric scale, or canonical frames. Building on these contact priors, S3AL pairs contact-aware optimization with fixed-stage VLM verification to annotate over 1M in-the-wild frames, enabling mesh-free, object-centric 3D learning of canonical pose, extent, refined hands, and signed distances from a single image.
\section{S3AL: A Scalable HOI Data Forge}
\label{sec:annotation}

\begin{figure*}[t]
  \centering
  \includegraphics[width=\textwidth]{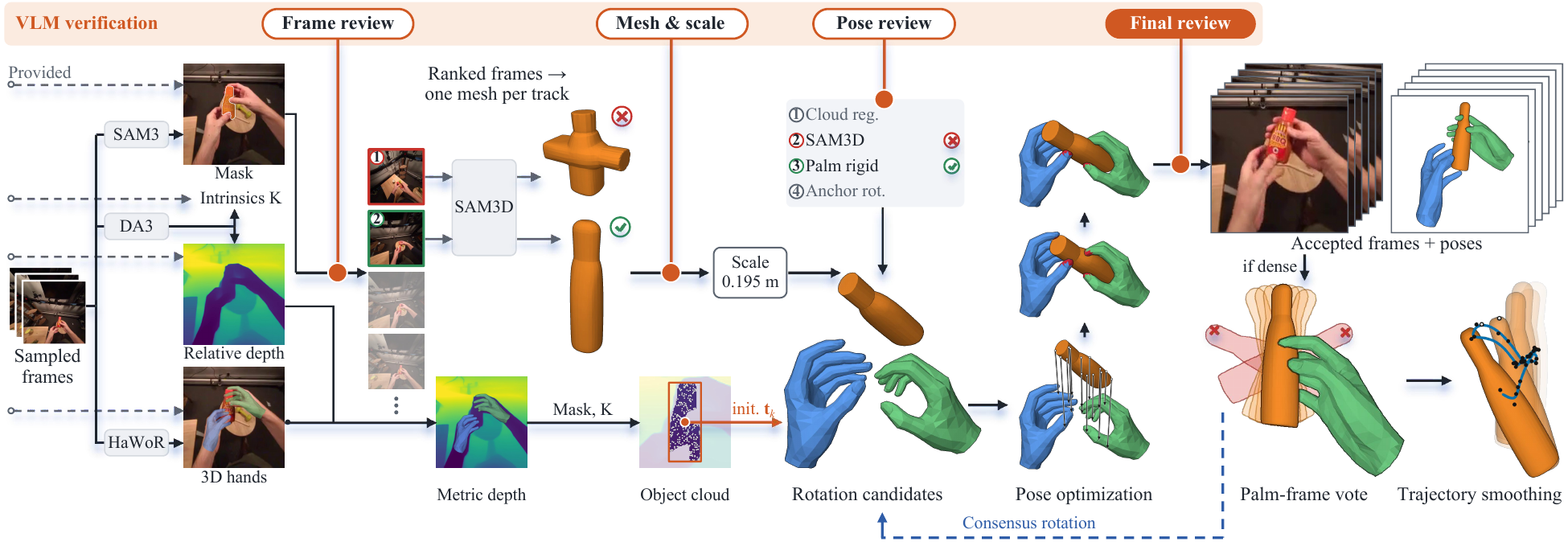}
  \caption{\textbf{Overview of S3AL.} 
  We leverage pretrained priors to provide masks, hand-calibrated metric depth, per-track canonical meshes, and rotation proposals. 
  S3AL decouples four VLM stages (orange) for bounded-cost semantic selection and filtering from two-stage ray-and-contact pose optimization, followed by palm-frame voting and trajectory smoothing.}
  \label{fig:annotation}
\end{figure*}

In this section, we introduce our 3D HOI pipeline S3AL with the overview presented in \Cref{fig:annotation}.
For more details, please refer to \cref{sec:supp_pipeline,sec:supp_prompts}.

\subsection{Pre-Processing}
\label{sec:preprocess}

Given a human video, we sample keyframes at a fixed rate and use off-the-shelf approaches to estimate target masks~\cite{carion2025sam3segmentconcepts}, camera-frame hand poses~\cite{zhang2025hawor}, relative depth $z^{\text{rel}}$, and camera intrinsics $K$.
Each object trajectory in the video is represented by a canonical mesh with per-frame object-to-camera rotation $R_k\in SO(3)$ and metric translation $\mathbf{t}_k\in\mathbb{R}^3$, mapping vertices $\mathbf{v}\mapsto R_k\mathbf{v}+\mathbf{t}_k$.
To recover metric depth from monocular predictions, we calibrate depth predicted by Depth Anything 3 (DA3)~\cite{depthanything3} against the metric MANO hands.
Specifically, we estimate the depth scale:
$s_h =
\bar{z}^{h}_{\Omega_h} / \bar{z}^{\text{rel}}_{\Omega_h}$,
where $\bar{z}^{h}_{\Omega_h}$ and $\bar{z}^{\text{rel}}_{\Omega_h}$ denote the mean hand and relative depths over valid pixels $\Omega_h$, respectively.
Back-projecting the object mask using $s_h z^{\text{rel}}$ yields a metric point cloud, whose centroid initializes $\mathbf{t}_k$.
In \emph{frame review}, given object mask overlays of sampled keyframes, a VLM verifies mask quality, maintains target identity across the trajectory, and identifies hand-object contact frames for subsequent pose estimation. 
It also ranks candidate anchors for 3D mesh reconstruction, favoring minimally occluded views regardless of contact.
From the highest-ranked anchor frame $a$, SAM3D~\cite{Chen_2026_CVPR} reconstructs a canonical mesh and initial orientation $R_a$. 
During \emph{mesh and scale review}, the VLM validates the mesh against multi-view renders and falls back to the next-ranked anchor if rejected.
Using the RGB scene, anchor crop, hand-calibrated depth map with statistics, and orthographic mesh-axis renders, the VLM estimates the object's longest metric dimension from visual depth cues and commonsense size priors to initialize uniform scaling.
It then refines axis-wise scales where object extents are visually grounded while preserving raw mesh proportions on occluded axes.

\subsection{Contact-Guided Per-Frame Pose Recovery}
\label{sec:pose}
\noindent\textbf{Rotation Proposals.}
For each contact frame $k$, we generate four rotation candidates in priority order: (1) 3D registration of depth-lifted 2D keypoint matches within the mask from anchor $a$ to frame $k$, (2) SAM3D's per-frame orientation estimate on frame $k$ (reusing the canonical mesh), (3) rigid propagation with the palm from anchor $a$, and (4) the static anchor rotation $R_a$. 
Palm propagation follows the interacting hand $h$ via
\begin{equation}
  R_k=\Delta R^{h}_{a\to k}\,R_a,
  \label{eq:palm}
\end{equation}
where $\Delta R^{h}_{a\to k}$ is the Kabsch rotation~\cite{https://doi.org/10.1107/S0567739476001873} aligning the wrist and knuckle joints of hand $h$ from frame $a$ to frame $k$, applied only when anchor $a$ is in contact with $h$ and $h$ maintains contact across $[a,k]$. 
In the \emph{pose review}, the VLM inspects rendered hand-object overlays in priority order, selecting the first plausible rotation $R_k$ or rejecting the frame if none aligns with the image.

\noindent\textbf{Pose Refinement and Verification.}
Monocular depth centroids misplace objects by $5$--$15$\,cm along viewing rays, where 2D silhouettes lack depth cues and heuristic 3D proximity losses fail under initial drift. 
With mesh and scale fixed, we refine poses in two stages.

In Stage~1, holding $R_k$ fixed, we align the projected mesh to the 2D mask center and cast camera rays through pixels where the projected hand and object meshes overlap. 
To anchor viewing-ray depth against the metric hand, we retain palm-facing ray hits ($\mathbf{x}_r^{\text{hand}},\mathbf{x}_r^{\text{obj}}$) whose front-to-back order matches mask visibility. 
With ray offset $\mathbf{o}_r=\mathbf{x}_r^{\text{hand}}-\mathbf{x}_r^{\text{obj}}$, outward hand normal $\mathbf{n}_r$, and per-hand-balanced weight $\alpha_r$, damped Newton steps update $\mathbf{t}_k\leftarrow\mathbf{t}_k+\Delta\mathbf{t}$ by minimizing
\begin{equation}
  E_{\text{ray}}=\sum_r \alpha_r\!\left(\lVert\Delta\mathbf{t}-\mathbf{o}_r\rVert_\epsilon
  +c\big[(\mathbf{o}_r-\Delta\mathbf{t})^{\top}\mathbf{n}_r\big]_+^2\right)\!,
  \label{eq:ray}
\end{equation}
where $\lVert\mathbf{y}\rVert_\epsilon=\sqrt{\lVert\mathbf{y}\rVert^2+\epsilon^2}$, $[s]_+=\max(0,s)$, and $c>0$ penalizes hand-object penetration.

In Stage~2, because Stage~1 holds $R_k$ fixed and aligns only coarse ray overlap, we resolve millimeter-scale finger-surface contact using pretrained TactoHand~\cite{xi2026tactohandlearninggeometrybaseddense}, which infers contact probabilities $p_i$ and target distances $\hat d_i$ from hand appearance alone for interacting-hand vertices $i\in\mathcal{V}$. 
Over active contacts $\mathcal{C}=\{i\in\mathcal{V}:p_i\ge 0.5\}$, coarse-to-fine damped Gauss--Newton refines $\mathbf{t}_k$ and a bounded rotation ($\le 20^\circ$) around the centroid, minimizing
\begin{equation}
  E_{\text{con}}=\frac{1}{2}\sum_{i\in\mathcal{C}}\!\Big[\delta\tanh\!\Big(\frac{d_i-\hat d_i}{\delta}\Big)\Big]^2
  +\sum_{i\in\mathcal{V}}\big[-d_i\big]_+^2,
  \label{eq:contact}
\end{equation}
over signed mesh distances $d_i(R_k,\mathbf{t}_k)$ ($<0$ inside) with margin $\delta$. 
An occlusion-aware mask IoU gate backtracks steps that degrade silhouette overlap. During \emph{final review}, the VLM inspects multi-view renders and retains only poses preserving the observed interaction without gross error.

\subsection{Temporal Consolidation}
\label{sec:dense}
When keyframes are sampled densely (\eg, 8\,FPS) across a contact segment, we enforce temporal consistency in two steps. 
First, we map first-pass accepted rotations into the rigid palm frame (\cref{eq:palm}) to vote for a consensus rotation and re-run pose review, optimization, and final review with this consensus as top candidate, eliminating symmetric flips. 
Second, to recover full-rate trajectories, we initialize intermediate source frames via pose interpolation and smooth the full sequence by solving
\begin{equation}
  \min_{\{R_k,\mathbf{t}_k\}}\;\sum_k w_k\,\rho_\tau\!\big(r_k\big)
  +\lambda\sum_k\big\lVert\ddot{\mathbf{x}}_k\big\rVert^2,
  \label{eq:smooth}
\end{equation}
where $r_k$ is the RMS distance between surface points under $(R_k,\mathbf{t}_k)$ and the per-frame estimate, $w_k$ downweights unaccepted and interpolated frames, $\rho_\tau$ is the Huber loss, and $\ddot{\mathbf{x}}_k$ is the surface-point acceleration.

\noindent\textbf{Scaling In-the-Wild Annotation.}
To augment hand-only tracks from a large-scale egocentric corpus~\cite{egodex} with 3D object geometry and contact, we run S3AL across all 168k episodes at 1\,FPS and densify selected segments to 8\,FPS using each track's canonical mesh. 
Consecutive accepted keyframes are smoothed into continuous trajectories while isolated ones remain single-frame labels, yielding over 1M in-the-wild frames with only a 0.2\% gross error rate across 5{,}000 audited samples. 
To supply GT MANO mesh fits (absent in this corpus) and standardized evaluation splits, we also convert contact-filtered studio captures into the same canonical representation.

\section{Learning Mesh-Free Object Abstractions}
\label{sec:abstraction}

\begin{figure*}[t]
  \centering
  \includegraphics[width=\textwidth]{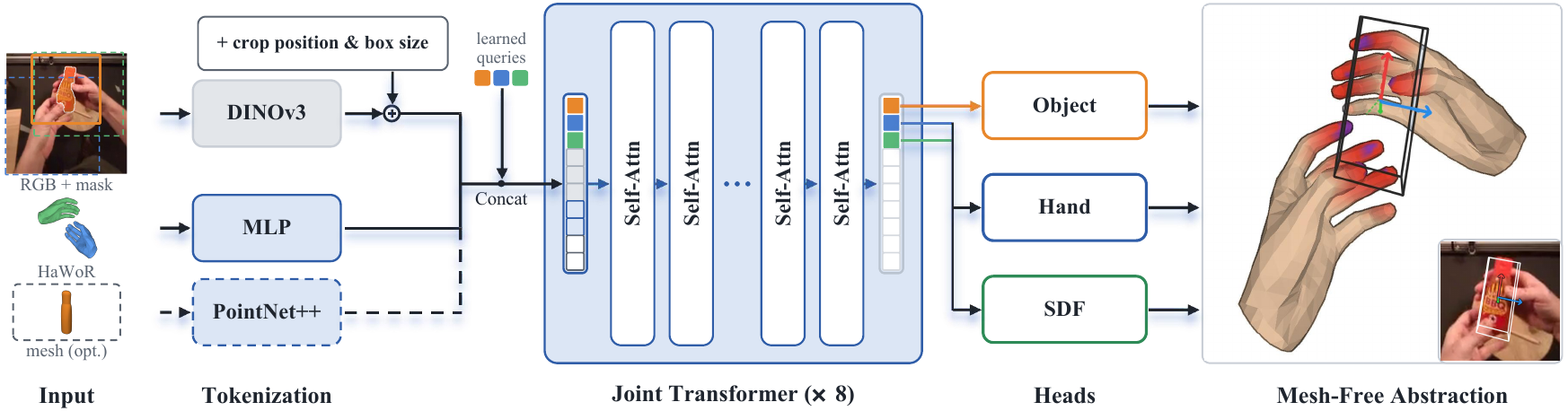}
  \caption{\textbf{Architecture of the mesh-free abstraction model.} 
  Multi-crop and full-frame DINOv3 tokens (frozen, gray) with crop geometry embeddings, HaWoR hand tokens, and optional PointNet++ mesh tokens are processed jointly with three learned queries by an 8-layer self-attention transformer (trainable, blue) to predict the canonical frame and box, refined hands, and signed distances.}
  \label{fig:model}
\end{figure*}

For mesh-free, object-centric 3D representation learning without decoding meshes, we train a single-pass predictor (\cref{fig:model}) to infer a compact \emph{mesh-free object abstraction}. 
Given an RGB image $I$ with intrinsics $K$, an object mask, and off-the-shelf MANO hands (plus an optional reference mesh when available), three heads output
\begin{enumerate*}[label=(\roman*)]
  \item a canonical object frame $P=(R,\mathbf{t})\in SE(3)$ in metric camera space with extent
        $\mathbf{e}\in\mathbb{R}^3_{+}$,
  \item refined MANO hands expressed in that frame, and
  \item signed distances from all 778 MANO vertices and 21 joints to the object.
\end{enumerate*}

\subsection{Object-Centric Representation}
\label{sec:abs_def}

\begin{figure}[b]
  \centering
  \begin{subfigure}[b]{0.49\linewidth}
    \centering
    \includegraphics[width=0.94\linewidth]{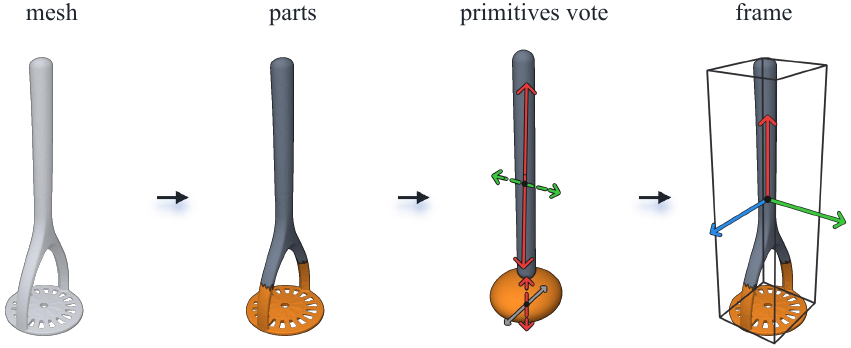}
    \caption{Frame rule}
    \label{fig:frame_rule}
  \end{subfigure}\hfill
  \begin{subfigure}[b]{0.49\linewidth}
    \centering
    \includegraphics[width=0.94\linewidth]{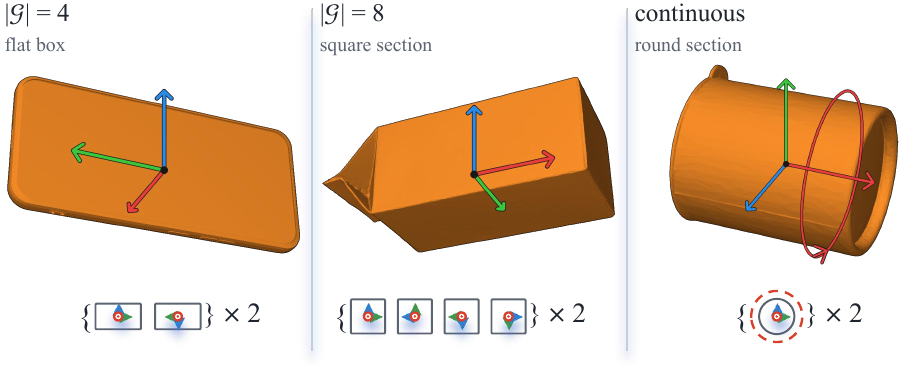}
    \caption{Label sets}
    \label{fig:frame_set}
  \end{subfigure}
  \caption{\textbf{Geometry-defined canonical frame.} \textbf{(a)} Decomposing a mesh into parts whose fitted primitives vote for the canonical axes (red, green, blue) and tight bounding box. \textbf{(b)} Geometrically indistinguishable orientations under discrete ($|\mathcal{G}|{=}4,8$) or continuous axial symmetry form the equivalence set $\mathcal{G}$.}
  \label{fig:frame}
\end{figure}

\noindent\textbf{Canonical Frame Construction.}
We derive the canonical frame purely from object geometry (\cref{fig:frame_rule}). 
Specifically, we partition sampled mesh surface points into up to three parts via iterative capsule clustering (adding a part only if it halves the primitive fit residual) and fit an oriented bounding box to each part. 
Each part casts votes for its longest and shortest box axes, weighted by its point share and axis elongation. 
Clustering these votes yields the primary axis from the strongest cluster and the secondary axis from the strongest orthogonal cluster, with point-projection asymmetry (skewness) fixing axis signs. 
When the two clusters stem from distinct parts (\eg, mug body and handle), the origin is placed at the midpoint of the shortest segment between the two part-centered axes, and otherwise at the main part's center. 
Since this origin may differ from the box center, each axis's percentile-trimmed projection range sets the extent $\mathbf{e}$ (span) and box center (midpoint).

\noindent\textbf{Symmetry-Aware Frame Labels.}
Object geometry does not always uniquely determine a single canonical frame: candidate axes may receive comparable vote scores, square (or near-square) cross-sections exhibit $90^\circ$ rotational symmetry, and near-symmetric shapes have ambiguous axis signs. 
We retain all valid orientations as a discrete symmetry label set $\mathcal{G}=\{F_m\}\subset SO(3)$ (\cref{fig:frame_set}). 
For continuous symmetries that a discrete set cannot capture, two binary flags mark objects as isotropic (\eg, spheres lacking a dominant axis) or continuously symmetric about the primary axis (\eg, circular cross-sections).

\noindent\textbf{Hand and Geometric Supervision.}
Where ground-truth MANO fits are available, we transform them into the canonical object frame as hand targets; across all sources, we precompute the signed distance $d$ from each reference MANO vertex and joint (GT in studio data, HaWoR on in-the-wild data) to the object mesh, clipped to $[-2,5]$\,cm, plus binary vertex contact labels ($|d|<5$\,mm).

\subsection{Model Design}
\label{sec:abs_net}

As shown in \cref{fig:model}, a frozen DINOv3 ViT-L/16~\cite{simeoni2025dinov3} encodes both the resized full-frame image $I$ for global context and local crops of the object (localized by its mask) and both hands. 
Because local cropping discards perspective location and scale~\cite{li2022cliff}, we add patch camera-ray and crop-scale embeddings to every visual token. For each input HaWoR~\cite{zhang2025hawor} hand, an MLP encodes per-joint features and global MANO parameters into tokens, and when an optional mesh is provided, we randomly rotate it to prevent canonical-frame leakage before encoding its surface points and metric size with PointNet++~\cite{qi2017pointnetplusplus}. 
All input tokens are concatenated with three learned queries (one for the object and one per hand) and processed jointly by an 8-layer (18.7M-parameter) transformer trunk, running in 33\,ms per image (including DINOv3) on an RTX~4090.

From the output object query, the object head regresses a 6D rotation~\cite{Zhou_2019_CVPR} relative to the crop-center ray, the frame origin (as a 2D pixel offset and log depth), log extents and box center, and two continuous-symmetry logits. 
The hand head predicts zero-initialized camera-space MANO residuals, expressing the refined hands in the predicted object frame via $\mathbf{x}^{\text{obj}}=R^{\top}(\mathbf{x}^{\text{cam}}-\mathbf{t})$. 
Finally, the SDF head regresses vertex and joint signed distances and vertex contact logits directly from each hand query, decoupling contact estimation from object-pose error.

\subsection{Training Objective}
\label{sec:abs_loss}

Let $\mathcal{P}_{24}$ denote the 24 proper signed permutation matrices ($\det=+1$) that relabel principal axes. 
For a predicted rotation $R$, we find the closest label frame up to axis relabeling,
\begin{equation}
  S_{\text{any}}=\argmax_{S\in\{Fg\,:\,F\in\mathcal{G},\,g\in\mathcal{P}_{24}\}}\operatorname{tr}(S^{\top}R),
  \label{eq:axisset}
\end{equation}
and the closest valid orientation $S_{\text{valid}}$ without relabeling over $\mathcal{G}$ (extended by closed-form axial rotations under continuous symmetry). 
Our \emph{axis-set loss},
\begin{equation}
  \mathcal{L}_{\text{rot}}=\lVert R-S_{\text{any}}\rVert_F+\mu\,\lVert R-S_{\text{valid}}\rVert_F,
  \label{eq:rotloss}
\end{equation}
accepts any valid alignment of principal axes, whose order is often ambiguous under occlusion, while softly favoring the default order via $\mu$; isotropic objects incur zero loss.

Axis-dependent targets ($\mathbf{e}$ and box center) are permuted to match $S_{\text{any}}$. 
To match the canonical-frame hand targets without leaking object-pose errors, hand losses transform predicted camera-space hands into the GT (rather than predicted) object frame. 
Standard regression and classification losses supervise the origin, extent, symmetry flags, hand parameters, signed distances, and contact, omitting MANO parameter losses on in-the-wild data, which lacks GT MANO fits.
\section{Experiments}
\label{sec:exp}

\begin{table}[b]
\centering
\caption{\textbf{Annotation accuracy and per-frame cost} on 30\,FPS HO3D v3 (estimated hands/masks; CD in cm). $^\dagger$GT mesh + DA3 depth. $^\ddagger$DA3 depth. $^\S$HO3D-trained. $^\P$GT hands. $^\ast$8.1\% SfM subset. EasyHOI: uncalibrated; AgentSTAR: subset.}
\label{tab:ann_main}
\footnotesize
\setlength{\tabcolsep}{3.4pt}
\begin{tabular*}{0.85\linewidth}{@{\extracolsep{\fill}}l ccc rr@{}}
\toprule
& \multicolumn{3}{c}{Accuracy} & \multicolumn{2}{c}{Cost / frame} \\
\cmidrule(lr){2-4}\cmidrule(l){5-6}
Method & CD$\downarrow$ & CD$^2\downarrow$ & F@10$\uparrow$ & GPU s & API \$ \\
\midrule
FoundationPose$^\dagger$~\cite{foundationposewen2024} & 3.12 & 67.5 & .360 & 0.23 & -- \\
Any6D$^\ddagger$~\cite{lee2025any6d} & 8.06 & 541.0 & .189 & 0.99 & -- \\
ForeHOI~\cite{chen2026forehoifeedforward3dobject} & 11.36 & 1072.5 & .164 & 0.83 & -- \\
HORT$^\S$~\cite{chen2025hort} & 3.33 & 36.3 & .239 & 0.17 & -- \\
DoAsIDo$^\P$~\cite{paliwal2026doasido} & 2.81 & 41.1 & .326 & $\sim$38 & -- \\
MagicHOI$^\ast$~\cite{wang2025magichoi} & 3.39 & 47.2 & .247 & $\sim$73 & -- \\
EasyHOI~\cite{liu2025easyhoi} & -- & -- & -- & $\sim$138 & -- \\
AgentSTAR~\cite{mazur2026agenticstar} & -- & -- & -- & 237 & 0.59 \\
\midrule
\textbf{S3AL} & \textbf{2.02} & \textbf{19.1} & \textbf{.456} & 4.03 & 0.0019 \\
\bottomrule
\end{tabular*}
\end{table}

Our experiments evaluate (i) whether S3AL recovers accurate, temporally coherent 3D HOI labels at scalable cost (\cref{sec:exp_ann,sec:exp_abl}), and (ii) whether our mesh-free abstraction is learnable from single images and generalizes in the wild (\cref{sec:exp_model}). 
We benchmark 3D reconstruction on egocentric HOT3D~\cite{banerjee2025hot3d} and third-person HO3D v3~\cite{hampali2020honnotate}, reporting symmetric Chamfer distance in metric camera space (CD, cm), its squared variant (CD$^2$, cm$^2$, which heavily penalizes gross failures~\cite{chen2023gsdf,chen2025hort}), F-score~\cite{8954121} at 10\,mm (F@10), and the fraction of frames below 5\,cm CD ($<$5).

We compare against pose trackers FoundationPose~\cite{foundationposewen2024} (with the GT mesh, FP+GT, or our SAM3D mesh) and Any6D~\cite{lee2025any6d}, monocular HOI methods ForeHOI~\cite{chen2026forehoifeedforward3dobject}, HORT~\cite{chen2025hort} (trained on HO3D and DexYCB), DoAsIDo~\cite{paliwal2026doasido} (using GT hands), MagicHOI~\cite{wang2025magichoi}, and EasyHOI~\cite{liu2025easyhoi}, and the agentic tracker AgentSTAR~\cite{mazur2026agenticstar}. 
All methods receive identical RGB images, masks, and intrinsics; FoundationPose and Any6D also receive DA3 depth calibrated to metric scale via GT hands. 
Protocol details and extended results (dense HOT3D, direct 3D box IoU, ADD/ADD-S~\cite{hinterstoisser2012model}) are in \cref{sec:supp_eval,sec:supp_results}.

\subsection{Comparison with SoTA Approaches}
\label{sec:exp_ann}

\begin{figure*}[t]
\centering
\includegraphics[width=0.85\linewidth]{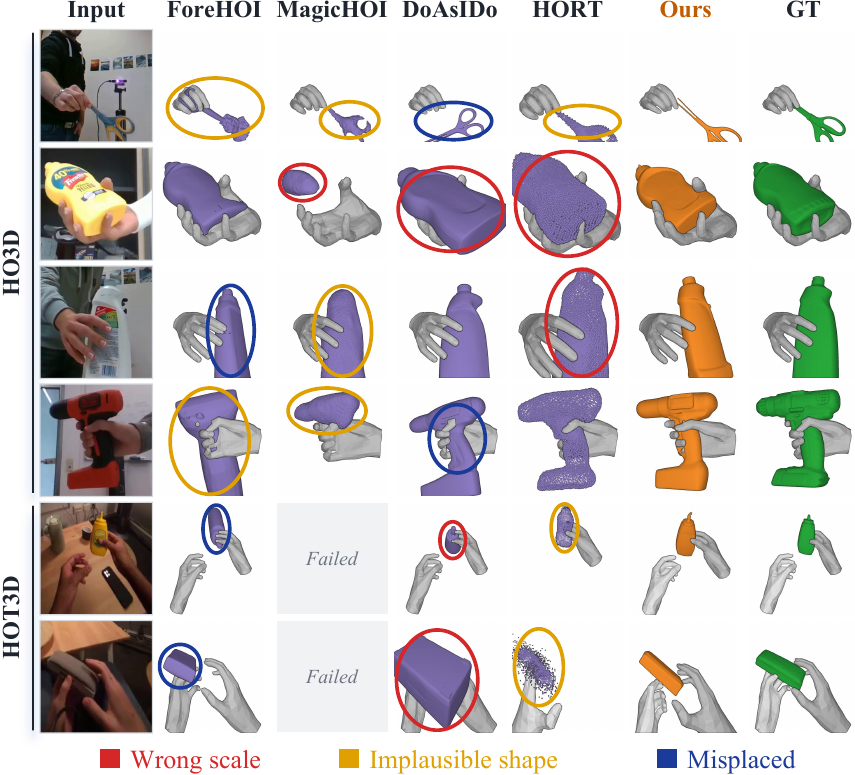}
\caption{\textbf{Qualitative annotation comparison on HO3D v3 and HOT3D} using estimated hands and masks (\emph{Failed}: SfM failure). Colored rings mark gross failures.}
\label{fig:ann_qual}
\end{figure*}

We annotate keyframes at 8\,FPS and smooth them to 30\,FPS, evaluating all baselines on contact frames kept by final review. 
On HO3D v3 (\cref{fig:ann_qual}), ForeHOI degrades out of domain, DoAsIDo suffers from pose drift and scale errors, MagicHOI yields truncated or underscaled meshes, and in-domain HORT distorts shape and scale. 
On HOT3D, MagicHOI fails COLMAP SfM (\emph{Failed}), while out-of-domain HORT misplaces or distorts unseen objects and misses bimanual grasps. 
S3AL instead recovers complete metric meshes firmly in grasp.
On HO3D with estimated hands and masks (\cref{tab:ann_main}), S3AL outperforms all baselines on the shared accepted frames across every metric, cutting CD$^2$ by 47\% over the runner-up (HORT). 
Similar gains hold on HOT3D and with GT hands, and trajectory smoothing cuts acceleration jitter by over $30\times$ on HOT3D and by an order of magnitude across pose baselines (\cref{tab:temporal}).

In \cref{tab:ann_main}, amortizing 8\,FPS keyframe compute and VLM calls across 30\,FPS HO3D trajectories on a single RTX~4090, S3AL achieves high accuracy at a fraction of prior cost. 
Feed-forward baselines run in under a second per frame but suffer frequent failures, while optimization pipelines run an order of magnitude slower, with MagicHOI failing COLMAP on most sequences and EasyHOI lacking metric scale. 
AgentSTAR needs dramatically higher GPU time and API cost with Claude Opus 5.5.

\begin{table}[t]
\centering
\caption{\textbf{Ablation of S3AL} on unsmoothed HO3D v3 keyframes at 8 FPS. Cov.: keyframe retention rate (\%); CD: cm; Cost: API cost per retained keyframe (USD). $^\ast$All outputs retained.}
\label{tab:ablation_ann}
\footnotesize
\setlength{\tabcolsep}{4.2pt}
\begin{tabular*}{0.85\linewidth}{@{\extracolsep{\fill}}lcccc@{}}
\toprule
Setting & Cov. & CD$\downarrow$ & CD$^2\downarrow$ & Cost (\$) \\
\midrule
\textbf{S3AL (full)} & 42.8 & \textbf{2.12} & \textbf{21.5} & 0.0068 \\
\midrule
w/o frame/pose review & 58.8 & 2.37 & 26.8 & 0.0026 \\
w/o final review & 47.9 & 2.26 & 24.9 & 0.0053 \\
w/o VLM$^\ast$ & 82.8 & 2.88 & 37.1 & 0 \\
\midrule
w/o contact optimization & 37.2 & 2.35 & 26.2 & 0.0084 \\
\bottomrule
\end{tabular*}
\end{table}

\subsection{Ablation Study of S3AL}
\label{sec:exp_abl}
\Cref{tab:ablation_ann} ablates the VLM verification stages and contact optimization on unsmoothed 8\,FPS HO3D keyframes. 
Dropping all VLM stages inflates outlier-sensitive CD$^2$ by 73\% (and $5\times$ under oracle inputs in \cref{tab:supp_vlm}), showing that monocular symmetries and mesh defects persist even without upstream detection noise.
Skipping final review retains post-optimization outliers for minimal coverage gain, while replacing upstream frame and pose reviews with heuristics lowers raw API spend but feeds poor rotation candidates into optimization.
Notably, on the same unscreened 30\,FPS frames (\cref{tab:supp_novlm}), \emph{S3AL w/o VLM} (2.79\,cm CD) still outperforms every baseline, showing that our gains stem from geometric reconstruction rather than VLM cherry-picking.
Crucially, removing contact optimization degrades 3D accuracy, lowers VLM acceptance, and raises effective API cost per retained keyframe by 24\%: because the VLM only filters discrete hypotheses, continuous contact refinement is needed to align candidates into valid, verifiable contact.

\subsection{Analysis of Mesh-Free Abstraction}
\label{sec:exp_model}

We train our predictor on a balanced 2\,FPS egocentric set of 0.4M samples, combining S3AL in-the-wild annotations (over half the data) with standardized studio sources supplying GT MANO fits, and evaluate on held-out splits, in-domain HOT3D, and zero-shot third-person HO3D v3. 
To compare with mesh-based baselines on surface CD (with direct mesh-free 3D box and ADD/ADD-S evaluation in \cref{tab:supp_model_full}), metric meshes (track-scaled SAM3D or GT) are placed in the predicted frame via the signed axis permutation maximizing 2D mask IoU.

\begin{figure}[b]
\captionsetup{justification=raggedright,singlelinecheck=false}
\begin{minipage}[t]{0.57\textwidth}
\centering
\includegraphics[width=\linewidth]{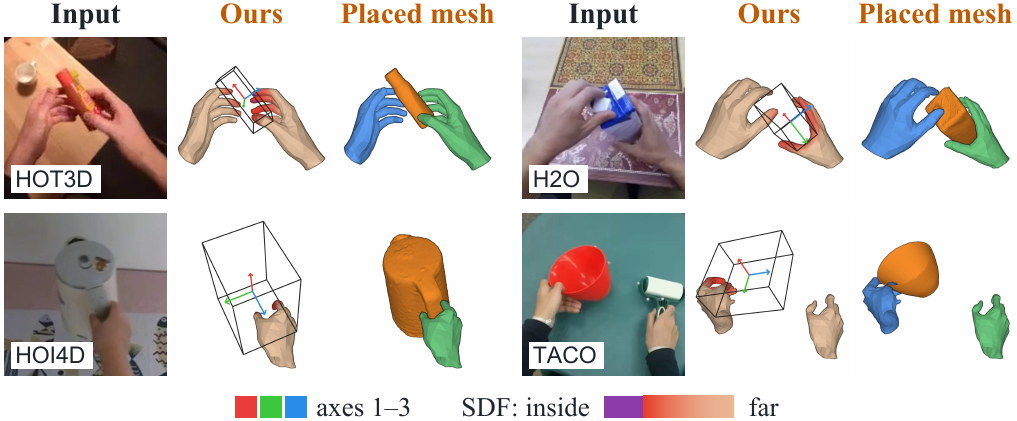}
\caption{\textbf{Single-image abstraction predictions across held-out test episodes} (median-error frames). \emph{Ours}: predicted oriented box, canonical axes, and refined hands colored by vertex signed distance. \emph{Placed mesh}: GT mesh aligned to our predicted frame.}
\label{fig:qual}
\end{minipage}\hfill
\input{fig/epic_wild}
\end{figure}

\noindent\textbf{Main Results.}
In \cref{fig:qual}, our model accurately predicts canonical axes, tight oriented bounding boxes, refined hand poses, and per-vertex signed distances across diverse objects, allowing placed meshes to align naturally within the grasp. 
Guided by the tie-breaking weight $\mu$ (\cref{eq:rotloss}), the network also converges to a consistent axis order and sign.
On in-domain HOT3D (\cref{tab:model}), given only RGB, masks, and estimated hands, our model outperforms all shape-reconstructing baselines without mesh input, remaining reliable under heavy hand occlusion where render-and-compare tracking fails.
On zero-shot third-person HO3D v3, it surpasses FoundationPose in CD and CD$^2$ despite the egocentric-to-third-person depth shift.
Supplying optional reference-mesh tokens resolves this cross-view scale-depth ambiguity and cuts CD from 4.28 to 2.40\,cm, outperforming even in-domain HORT and GT-hand DoAsIDo across all metrics. 
Joint object-hand reasoning also cuts camera-space hand MPJPE by 63\% over HaWoR and outperforms InterField~\cite{fan2023arctic} on signed distances (\cref{tab:hands}), confirming that our fixed-size abstraction is well-conditioned and readily learnable in a single forward pass.

\begin{table*}[t]
\centering
\caption{\textbf{Single-image 3D object placement} on HOT3D and HO3D v3 (CD in cm). \emph{Shape} denotes the evaluated 3D geometry (\emph{Self-recon.}: reconstructed by the method itself). $^\P$Uses GT hands. $^\S$Trained on HO3D.}
\label{tab:model}
\footnotesize
\setlength{\tabcolsep}{5.2pt}
\begin{tabular}{@{}llc cccc c cccc@{}}
\toprule
& & & \multicolumn{4}{c}{HOT3D (in-domain)} & & \multicolumn{4}{c}{HO3D v3 (zero-shot)} \\
\cmidrule(lr){4-7}\cmidrule(l){9-12}
Method & Shape & Depth & CD$\downarrow$ & CD$^2\downarrow$ & F@10$\uparrow$ & $<$5$\uparrow$ & & CD$\downarrow$ & CD$^2\downarrow$ & F@10$\uparrow$ & $<$5$\uparrow$ \\
\midrule
FoundationPose~\cite{foundationposewen2024} & SAM3D & \checkmark & 6.12 & 176.0 & .196 & 56.7 & & 5.47 & 162.6 & .236 & 70.7 \\
Any6D~\cite{lee2025any6d} & Self-recon. & \checkmark & 7.67 & 435.0 & .153 & 50.6 & & 8.80 & 447.6 & .150 & 46.1 \\
ForeHOI~\cite{chen2026forehoifeedforward3dobject} & Self-recon. & -- & 9.60 & 629.6 & .137 & 42.3 & & 15.63 & 1930.4 & .114 & 36.7 \\
DoAsIDo$^\P$~\cite{paliwal2026doasido} & Self-recon. & -- & 5.24 & 114.5 & .192 & 61.3 & & 3.81 & 71.9 & .262 & 81.4 \\
HORT$^\S$~\cite{chen2025hort} & Self-recon. & -- & 4.24 & 90.8 & .214 & 71.3 & & 3.29 & 36.9 & .249 & 79.7 \\
\textbf{Ours} & SAM3D & -- & \textbf{1.95} & \textbf{19.8} & \textbf{.456} & \textbf{93.7} & & 4.28 & 64.8 & .227 & 64.5 \\
\textbf{Ours + mesh input} & SAM3D & -- & 2.29 & 26.2 & .425 & 90.5 & & \textbf{2.40} & \textbf{24.1} & \textbf{.370} & \textbf{91.0} \\
\midrule
FoundationPose & GT & \checkmark & 6.61 & 419.8 & .219 & 59.2 & & 5.13 & 140.9 & .248 & 69.8 \\
\textbf{Ours} & GT & -- & \textbf{1.72} & \textbf{16.7} & .512 & \textbf{94.5} & & 4.26 & 63.7 & .220 & 64.4 \\
\textbf{Ours + mesh input} & GT & -- & 1.81 & 20.5 & \textbf{.531} & 92.5 & & \textbf{2.59} & \textbf{27.0} & \textbf{.338} & \textbf{88.9} \\
\bottomrule
\end{tabular}
\end{table*}

\noindent\textbf{Analysis.}
\Cref{tab:ablation} ablates our model's key design choices.
Removing signed-distance supervision hurts contact alignment and degrades zero-shot HO3D generalization from 4.28 to 4.99\,cm CD.
Further dropping the HaWoR hand prior and residual parameterization nearly doubles camera-space MPJPE (27.1 to 47.6\,mm) while degrading object placement via weaker metric cues.
Finally, replacing our axis-set loss with $S_{\text{valid}}$-only rotation supervision forces ambiguous axis ordering and worsens every metric, whereas supplying optional mesh tokens further lowers error.
Adding S3AL supervision to studio training matches studio-level accuracy on unseen in-the-wild tasks and markedly improves zero-shot stability under clutter and occlusion on EPIC-KITCHENS-100~\cite{damen2022rescaling} (\cref{fig:epic_wild}).

\begin{table}[h!]
\centering
\caption{\textbf{Abstraction model ablation} on the test split and HO3D (CD in cm; MPJPE in mm; Obj.\ MPJPE in object frames; HO3D: SAM3D). $-$: cumulative removal. $^\dagger S_{\text{valid}}$ match.}
\label{tab:ablation}
\footnotesize
\setlength{\tabcolsep}{2.5pt}
\begin{tabular*}{0.85\linewidth}{@{\extracolsep{\fill}}lcccc c@{}}
\toprule
& \multicolumn{4}{c}{Test set} & HO3D \\
\cmidrule(lr){2-5}\cmidrule(l){6-6}
Model & CD$\downarrow$ & F@10$\uparrow$ & MPJPE$\downarrow$ & Obj.\ MPJPE$\downarrow$ & CD$\downarrow$ \\
\midrule
\textbf{Ours (full)} & \textbf{1.19} & \textbf{.631} & \textbf{26.6} & \textbf{47.1} & \textbf{4.28} \\
$-$ SDF supervision & 1.22 & .622 & 27.1 & 48.1 & 4.99 \\
$-$ HaWoR hand prior & 1.28 & .600 & 47.6 & 54.1 & 5.41 \\
$-$ axis-set loss ($S_{\text{valid}}$) & 1.31 & .590 & 51.9 & 65.3$^\dagger$ & 5.94 \\
\midrule
\emph{Ours + mesh input} & 1.09 & .659 & 26.4 & 44.9 & 2.40 \\
\bottomrule
\end{tabular*}
\end{table}

\FloatBarrier
\section{Conclusion}
\label{sec:conclusion}
In this paper, we present S3AL, a data forge to scale in-the-wild 3D HOI annotation for mesh-free learning.
Pairing fixed-stage VLM verification with contact-guided keyframe refinement and temporal smoothing, S3AL recovers accurate 3D trajectories with an order of magnitude less GPU time per frame than per-sequence 3D fitting and over $100\times$ lower API cost than agentic VLM trackers, curating over 1M reviewed frames.
Distilling these labels into a canonical object frame, extent, refined hands, and signed distances enables a feed-forward predictor to learn metric 3D alignment without mesh reconstruction, outperforming depth-based baselines and transferring to unseen objects and views.
This provides compact 3D supervision, with future work addressing articulated objects and fast occluded motion.
\section*{LLM Usage Statement}
As part of the proposed method and experimental evaluation (\cref{sec:annotation,sec:exp}), our annotation pipeline invokes Gemini 3.8 Flash as a fixed-stage vision-language verifier, and we evaluate the AgentSTAR baseline with Claude Opus 5.5 as well as GPT-6 Luna and Qwen 3.8 Flash as alternative VLM auditors in the appendix. During research and manuscript preparation, we also used Claude, Gemini, and ChatGPT to assist with implementation utilities, experiment-processing scripts, prose polishing, and \LaTeX{} formatting. All research ideas, methodological designs, experimental analyses, citations, and scientific claims were conceived, verified, and validated by the authors, who take full responsibility for the content of this paper.
\bibliographystyle{unsrtnat}
\bibliography{main}
\clearpage
\appendix
\noindent{\fontsize{17}{20}\selectfont\bfseries Appendix\par}
\vskip 3mm
\suppressfloats[t]
\crefalias{section}{appendix}
\newcommand{\suppimg}[2]{\includegraphics[width=#2]{fig/img/#1}}
\noindent This appendix details the S3AL data forge (\cref{sec:supp_pipeline}), our mesh-free object abstraction and predictor (\cref{sec:supp_model}), experimental protocols and evaluation metrics (\cref{sec:supp_eval}), extended quantitative and qualitative results (\cref{sec:supp_results}), limitations and future directions (\cref{sec:supp_limits}), and the four fixed-stage VLM verification prompts (\cref{sec:supp_prompts}).

\section{S3AL Data Forge Details}
\label{sec:supp_pipeline}

\begin{figure*}[t]
\centering
\suppimg{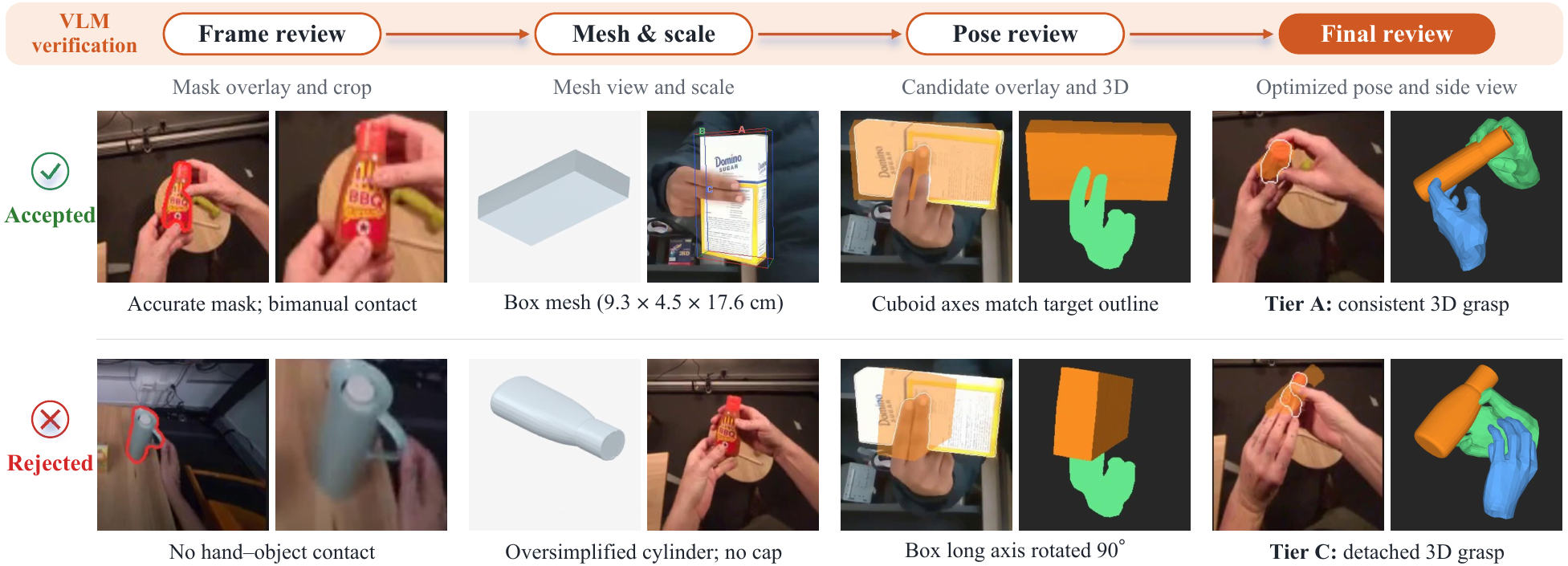}{\textwidth}
\caption{\textbf{VLM verification stages in action.} Each column shows one accepted and one rejected example from our full-auto HOT3D and HO3D runs, along with the key decision criteria.}
\label{fig:supp_checkpoints}
\vspace{-2mm}
\end{figure*}

This section follows the stage order of \cref{sec:annotation}. All four VLM verification stages (\cref{fig:supp_checkpoints}) call Gemini 3.8 Flash (\texttt{gemini-3.8-flash}) with schema-enforced JSON outputs at temperature $0$ (low thinking for frame, mesh, and scale review; medium thinking for pose and final review). 
Transient API or schema errors are retried rather than cached as semantic rejections, while self-contradictory or \emph{uncertain} responses are always rejected; \cref{sec:supp_prompts} summarizes the core prompt templates.

\subsection{Pre-Processing}
\noindent\textbf{Inputs and Metric Depth Calibration.}
Object masks come from SAM3~\cite{carion2025sam3segmentconcepts}, relative depth $z^{\text{rel}}$ and camera intrinsics $K$ from DA3~\cite{depthanything3}, and camera-space MANO hands from HaWoR~\cite{zhang2025hawor}. 
To estimate the hand-to-depth scale $s_h$ (\cref{sec:preprocess}), we render visible MANO hand depth $z_p^h$, compute the per-pixel ratio $r_p = z_p^h / z_p^{\text{rel}}$ over non-occluded hand pixels $p\in\Omega_h$, and retain the inlier pixel set $\mathcal{P}_{\text{in}}\subseteq\Omega_h$ lying within $3\,\text{MAD}$ of the median. The scale factor is then obtained via $s_h = \sum_{p \in \mathcal{P}_{\text{in}}} z_p^h \big/ \sum_{p \in \mathcal{P}_{\text{in}}} z_p^{\text{rel}}$.

Frames without a valid hand-depth scale are skipped.
For each valid frame $k$, we back-project object-mask pixels
$p=(u_p,v_p)$ with calibrated depth $z_p=s_h z_p^{\text{rel}}$
into camera coordinates as
$\mathbf{x}_p=z_p K^{-1}[u_p,v_p,1]^{\top}$.
Let $\Omega_k^{\text{obj}}$ denote the object-mask pixels
with valid depth. The initial object translation $\mathbf{t}_k$ is computed
as the centroid of the resulting point cloud:
\begin{equation}
\mathbf{t}_k^{(0)}
=
\frac{1}{|\Omega_k^{\text{obj}}|}
\sum_{p\in\Omega_k^{\text{obj}}} \mathbf{x}_p.
\label{eq:translation_initialization}
\end{equation}

\noindent\textbf{Frame Review.}
Frames with outlined candidate masks are evaluated in batches of eight. 
For each frame, the VLM checks 
\begin{enumerate*}[label=(\arabic*)]
    \item whether the mask is reliable for pose estimation,
    \item whether it tracks the same physical instance across the sequence, and
    \item whether the object view is complete enough for 3D reconstruction.
\end{enumerate*}
Meanwhile, it also classifies the interaction as \emph{contact}, \emph{no contact}, \emph{other object}, or \emph{uncertain}, and identifies the interacting hand(s). 
Only \emph{contact} frames passing these checks proceed to pose estimation.
Independently of contact state, all frames passing the three checks are ranked as mesh anchors, which allows minimally occluded, contact-free views to be selected.

\subsection{Canonical Mesh and Scale Estimation}
\noindent\textbf{Mesh Reconstruction and Review.}
SAM3D~\cite{Chen_2026_CVPR} reconstructs the canonical mesh and anchor orientation $R_a$ from the top-ranked anchor frame $a$. 
To verify each reconstructed SAM3D mesh, we prompt the VLM with a two-row composite image showing the anchor crop alongside six renders from opposing viewpoints. 
We discard meshes exhibiting missing parts, merged hand or background geometry, or an incorrect global shape, and fall back to the second-ranked anchor if the primary candidate fails.

\noindent\textbf{Metric Scale Estimation.}
Next, we query the VLM to estimate the metric dimensions of the object from the anchor image, object crop, hand-calibrated depth map, orthographic mesh renders, and projected mesh axes.
The model returns the longest overall dimension $m$ alongside per-axis extents, labeling each axis as \emph{supported} or \emph{unsupported}. 
Specifically, an axis is marked \emph{supported} when its axis correspondence and metric extent are visually grounded---explicitly including thin structures (\eg, plates, cards, or lids) whose reconstructed mesh thickness is artificially inflated.
Conversely, visually ungrounded axes are marked \emph{unsupported}, in which case the raw mesh proportions scaled by the longest dimension $m$ are preserved.
We first uniformly rescale the mesh to match the VLM's longest metric dimension, and then independently rescale each axis marked as supported to its estimated extent.

\subsection{Contact-Guided Per-Frame Pose Recovery}
\noindent\textbf{Rotation Proposals.}
\label{sec:supp_rot}
Single-view orientation estimation in the wild faces complementary failure modes: feature matching breaks on textureless objects or wide baseline shifts, per-frame 3D predictors can flip nearly symmetric shapes, and hand-driven propagation drifts when fingers re-grasp. For each contact frame $k$, we therefore construct four complementary rotation hypotheses in decreasing order of geometric specificity:
\begin{enumerate}
  \item \textbf{RGB-D correspondence registration.} SIFT keypoints matched inside the object masks of anchor frame $a$ and frame $k$ are lifted to metric 3D via calibrated DA3 depth. RANSAC with Kabsch alignment ($6$\,px reprojection and $2$\,cm 3D inlier thresholds) estimates the relative rigid motion $(\Delta R,\Delta\mathbf{t})$ and sets $R_k=\Delta R\,R_a$. Placed first because multi-view surface correspondences yield the sharpest alignment when texture is visible, and skipped automatically when inliers are insufficient.
  \item \textbf{Per-frame SAM3D orientation.} Because SAM3D~\cite{Chen_2026_CVPR} generates 3D assets in two stages---first predicting coarse structure and global 3D pose before decoding detailed surface geometry in stage~2---we run only lightweight stage-1 inference ($25$ steps, distillation disabled) on frame $k$, bypassing costly per-frame mesh synthesis while reusing the track's verified canonical mesh. Expressing this orientation relative to the anchor mesh recovers free 3D reorientation across large viewpoint changes or on textureless surfaces where 2D matching fails.
  \item \textbf{Palm-rigid propagation (\cref{eq:palm}).} Enabled only when the same hand $h$ maintains continuous contact across sampled frames between $a$ and $k$. To prevent finger articulation from corrupting object pose, $\Delta R^h_{a\to k}$ is fit by Kabsch alignment over only the five rigid palm joints (wrist and four MCP knuckles, MANO joint indices $\{0,5,9,13,17\}$), providing a reliable kinematic prior under heavy hand occlusion.
  \item \textbf{Static anchor fallback.} In many everyday manipulation segments (\eg, lifting, carrying, or linear placement), a held object translates with negligible rotational change; setting $R_k=R_a$ preserves the clean anchor orientation without injecting per-frame estimation drift when earlier proposals fail.
\end{enumerate}

\noindent\textbf{Pose Review.}
Evaluated in batches of four frames and two rotation candidates per round, each candidate displays the RGB crop, the rendered mesh overlay, and the rendered mesh in isolation. The VLM compares the principal axis, exposed face, and asymmetric features while separating rotation error from mesh-shape imperfection, returning \emph{accept}, \emph{reject}, or \emph{uncertain}. Round~1 tests priorities~(1)--(2); if neither is accepted, Round~2 tests priorities~(3)--(4), selecting the highest-priority accepted rotation or rejecting the frame.

\begin{figure}[t]
\centering
\suppimg{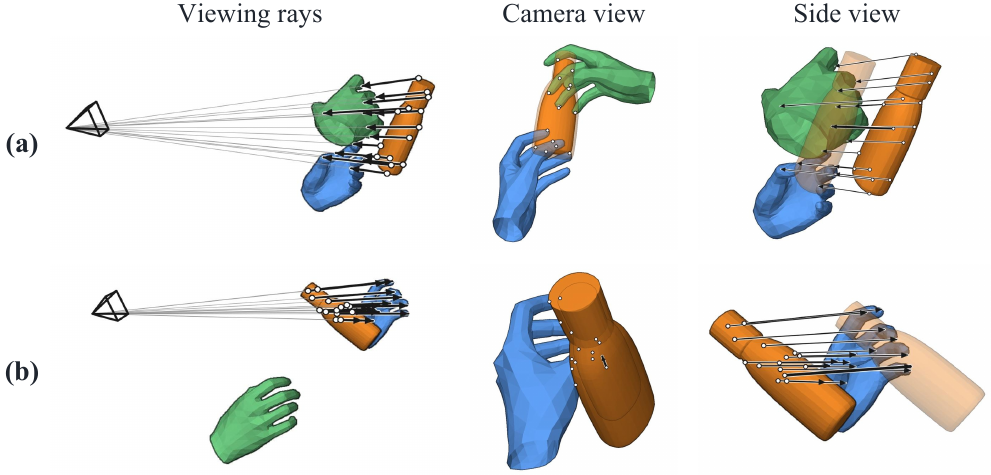}{0.75\linewidth}
\caption{\textbf{Stage 1: Ray-based translation.} From left to right: pinhole viewing rays with hand--object offsets $\mathbf{o}_r$, camera-view overlap before (solid) and after (translucent) optimization, and side-view shift $\Delta\mathbf{t}^\ast$. \textbf{(a)} Overestimated initial depth pulled $8.1$\,cm closer into a bimanual grasp. \textbf{(b)} Underestimated initial depth pushed $14.9$\,cm farther into the linked left hand while ignoring the unlinked right hand.}
\label{fig:supp_ray}
\vspace{-2mm}
\end{figure}

\noindent\textbf{Stage 1: Ray-Based Translation Optimization.}
On frames accepted by pose review, we hold the canonical mesh and metric scale fixed and refine $(R_k,\mathbf{t}_k)$ using only the hand(s) linked exclusively to the target object by frame review (all solver thresholds and weights below were fixed prior to full-scale annotation and never tuned on evaluation benchmarks). Monocular depth centroids frequently misplace the object by $5$--$15$\,cm along the viewing ray while keeping its 2D projection nearly unchanged (\cref{fig:supp_ray}). Holding the selected rotation $R_k$ fixed, we first align the 2D convex-hull center of the projected mesh to the mask center in $(x,y)$ at the median mesh-vertex depth (up to $12$ iterations, stopping below $0.05$\,px). Next, over pixels $\mathbf{u}$ where the projected object mesh and linked MANO hand(s) overlap in 2D, we cast unit pinhole rays $\mathbf{d}_{\mathbf{u}}=K^{-1}\tilde{\mathbf{u}}/\lVert K^{-1}\tilde{\mathbf{u}}\rVert$ from the camera origin and record first-hit points $\mathbf{x}_r^{\text{hand}}$ and $\mathbf{x}_r^{\text{obj}}$ (\cref{fig:supp_ray}). A ray $r$ is retained only if (i) the outward hand-surface normal $\mathbf{n}_r$ is palm-facing ($\mathbf{n}_r^{\top}\mathbf{n}_{\text{palm}}>-0.25$) and (ii) the front/back ray ordering of hand and object agrees with pixel mask visibility up to a $2$\,mm margin. Requiring at least two valid hits, we subsample up to $64$ rays per linked hand via 2D farthest-point sampling, forming ray--surface hit pairs $\mathcal{P}_h$ for each active hand $h\in\mathcal{H}$ with equal per-hand weight $\alpha_r=1/(|\mathcal{H}||\mathcal{P}_h|)$ and outward hand-triangle normal $\mathbf{n}_r$. With ray displacement $\mathbf{o}_r=\mathbf{x}_r^{\text{hand}}-\mathbf{x}_r^{\text{obj}}$ and smoothed norm $\lVert\mathbf{y}\rVert_\epsilon=\sqrt{\lVert\mathbf{y}\rVert^2+\epsilon^2}$ ($\epsilon=10^{-8}$\,m), we minimize \cref{eq:ray} ($c=20$\,m$^{-1}$) from the coordinate-wise median of $\{\mathbf{o}_r\}$ via damped Newton steps with Armijo backtracking line search ($\le 100$ iterations), shifting $\mathbf{t}_k\leftarrow\mathbf{t}_k+\Delta\mathbf{t}^\ast$; frames without a uniquely linked hand retain the hull-aligned translation.

\noindent\textbf{Stage 2: Contact-Guided Pose Refinement.}
Pretrained TactoHand~\cite{xi2026tactohandlearninggeometrybaseddense} processes $32$-frame windows of $256\times192$ hand crops ($\sim\!10$\,FPS) to predict per-vertex contact probability $p_i$ and proximity $q_i=\exp(-10\,d_i)$, decoded as $\hat d_i=-\log(\operatorname{clamp}(q_i,10^{-6},1))/10$\,m. With linked-hand vertex set $\mathcal{V}$ and active contact set $\mathcal{C}=\{i\in\mathcal{V}:p_i\ge 0.5\}$, frames with $|\mathcal{C}|\ge 1$ optimize the camera-space object centroid $\mathbf{c}$ ($\mathbf{t}_k=\mathbf{c}-R_k\bar{\mathbf{v}}$) and an RMS-radius-scaled rotation update projected within $20^\circ$ of Stage~1 (per-step caps of $1$\,cm and $2^\circ$) by minimizing \cref{eq:contact}, where $d_i$ is the signed distance to the largest mesh component (sign via five-ray occupancy majority voting). Damped Gauss--Newton ($\lambda=10^{-4}$, forward-mode autodiff Jacobians, truncated to singular values $\sigma\ge\max(10^{-7},10^{-3}\sigma_{\max})$) runs in two coarse-to-fine passes ($\delta=5$\,cm then $\delta=5$\,mm, $\le 100$ iterations each). At the end of each pass, we render the hand-occluded object silhouette ($512$\,px long side) and evaluate $s=1-\operatorname{IoU}_{\text{soft}}$ against the object mask; if $s>1.01\,s_0+10^{-6}$, we backtrack along geodesic/linear interpolation ($\alpha=2^{-m}$, $m=1,\dots,8$) and revert to the pass start if the gate stays violated or $E_{\text{con}}$ does not decrease. Frames with $|\mathcal{C}|=0$ retain the Stage~1 pose.

\noindent\textbf{Final Review.}
Before final review, a frame is rejected if ray-based translation finds fewer than two valid ray hits and the projected mesh covers less than $50\%$ of the object mask. Given the camera-view panels after pose optimization together with orthogonal side and top hand--object views, the VLM then assigns Tier~A (accurate grasp and object), Tier~B (valid grasp with minor shape, scale, or finger-detail discrepancy), Tier~C (wrong hand--object relation or visibly wrong rotation), or Tier~U (unjudgeable). Only Tiers~A and~B are retained.

\subsection{Temporal Consolidation}
\begin{figure*}[t]
\centering
\suppimg{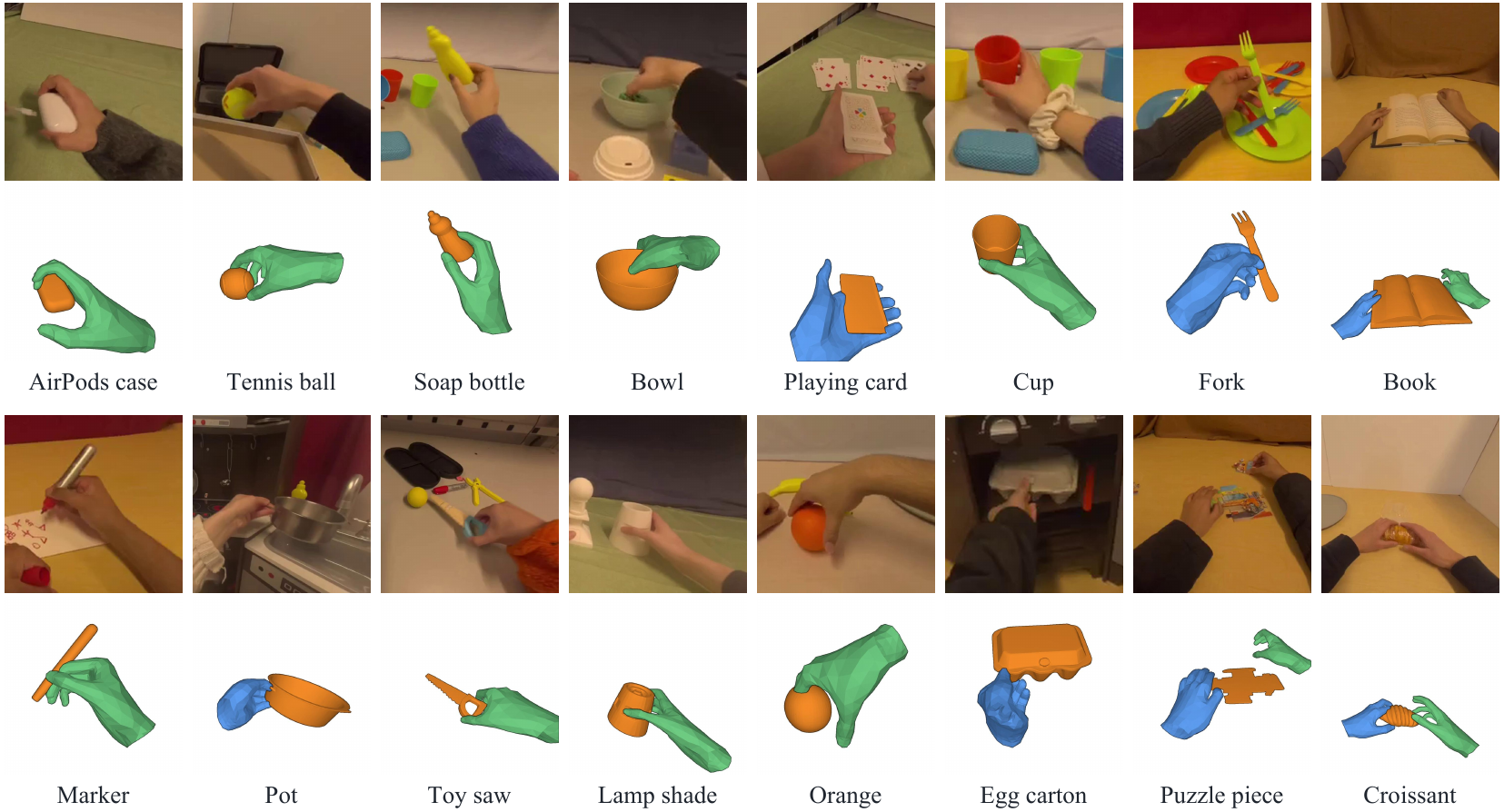}{\textwidth}
\caption{\textbf{Accepted in-the-wild 3D labels~\cite{egodex}.} Each object is shown in the input image and as the verified 3D label rendered from the same camera.}
\label{fig:supp_egodex}
\vspace{-2mm}
\end{figure*}

Keyframes are sampled at $8$\,FPS, reusing each track's accepted canonical mesh and metric scale. Across a continuous contact segment, a second pass adds a palm-frame consensus rotation as the top candidate before re-running pose review, optimization, and final review, after which keyframes are interpolated and smoothed to $30$\,FPS.

\noindent\textbf{Palm-Frame Consensus.}
Within a segment where the same hand maintains contact, accepted $8$\,FPS rotations are mapped into the rigid palm frame (\cref{eq:palm}). Each keyframe gathers votes within $\pm 30$ source frames; the rotation with the most inliers within $20^\circ$ wins, and the chordal mean of its inliers forms the top candidate (breaking two-hand ties by inlier count), which must still pass pose review.

\noindent\textbf{Interpolation and Trajectory Smoothing.}
Outputs separated by more than $15$ source frames ($0.5$\,s) start a new segment; missing frames inside a segment are initialized via rotation SLERP and linear translation interpolation (inheriting acceptance only when both bounding keyframes are accepted). Segments of at least three frames are smoothed by minimizing \cref{eq:smooth} in double precision with L-BFGS and strong Wolfe line search ($\le 1{,}000$ iterations), where $r_k$ is the RMS distance over $256$ area-uniform surface points, $\ddot{\mathbf{x}}_k$ is the second-order finite-difference acceleration over actual frame timestamps, $\tau=2$\,cm, $\lambda=(\sigma_{\text{obs}}/\sigma_{\text{acc}})^2$ ($\sigma_{\text{obs}}=1$\,cm, $\sigma_{\text{acc}}=2$\,m/s$^2$), and $w_k\in\{1,0.25,0\}$ for accepted, unaccepted, and interpolated frames, respectively.

\noindent\textbf{Density Filtering.}
Finally, an output keyframe is marked \emph{dense} if more than four outputs per second fall within $\pm 15$ source frames in its segment; interpolated frames are retained only between two dense outputs, whereas keyframe outputs themselves are always kept.

\noindent\textbf{Scaling In-the-Wild Annotation.}
We run S3AL across all 168k episodes of a large-scale in-the-wild egocentric corpus~\cite{egodex}, sampling keyframes at 1\,FPS across the full corpus and densifying selected manipulation segments to 8\,FPS while reusing each track's verified canonical mesh and metric scale. Consecutive accepted keyframes are smoothed into dense trajectories while isolated keyframes are retained as single-frame labels, yielding over 1M final-review-accepted in-the-wild 3D frames (\cref{fig:supp_egodex}). In a manual audit of 5{,}000 randomly sampled accepted frames---checking grasp association, 3D orientation, and metric scale---we observed a gross error rate of only 0.2\%.

\section{Mesh-Free Object Abstraction Details}
\label{sec:supp_model}

\subsection{Object-Centric Representation}
\label{sec:supp_frame}
Given an object mesh $\mathcal{M}$ in arbitrary coordinates, our deterministic geometry procedure operates without semantic category priors to extract a canonical orientation $R=[\mathbf{r}_1,\mathbf{r}_2,\mathbf{r}_3]\in SO(3)$, origin $\mathbf{o}\in\mathbb{R}^3$, percentile-trimmed extent $\mathbf{e}\in\mathbb{R}^3_+$, discrete symmetry set $\mathcal{G}\subset SO(3)$, and continuous-symmetry flags (\cref{fig:frame}). 
All thresholds below were fixed once prior to generating labels.

\noindent\textbf{Part Decomposition and Primitive Selection.}
To suppress sampling variance, we run five fixed random seeds with equal weights ($1/5$ each), sampling $4{,}000$ points uniformly over the mesh surface per seed and testing $k\in\{1,2,3\}$ parts via a capsule-shell variant of Lloyd's algorithm.
Each part $\mathcal{P}_j$ is parameterized as a capsule with centroid $\bar{\mathbf{x}}_j$, dominant PCA unit axis $\mathbf{a}_j$, and spine segment $\mathcal{S}_j=\{\bar{\mathbf{x}}_j+\alpha\mathbf{a}_j:\alpha\in[\ell_j^{\min},\ell_j^{\max}]\}$, whose endpoints $[\ell_j^{\min},\ell_j^{\max}]$ are clipped to the 2nd and 98th percentiles of the scalar projections $\{(\mathbf{x}-\bar{\mathbf{x}}_j)^{\top}\mathbf{a}_j\}_{\mathbf{x}\in\mathcal{P}_j}$. 
The capsule radius is defined as $r_j=\operatorname{median}_{\mathbf{x}\in\mathcal{P}_j}d(\mathbf{x},\mathcal{S}_j)$ using the Euclidean point-to-segment distance $d(\cdot,\cdot)$.
For a fixed $k$, we initialize the part centers via farthest-point sampling (starting from the point farthest from the global centroid) and perform up to $30$ iterations alternating between capsule refitting and point reassignment to the nearest capsule shell:
\begin{equation}
  j^\ast(\mathbf{x})=\argmin_{j\in\{1,\dots,k\}}\big|d(\mathbf{x},\mathcal{S}_j)-r_j\big|.
  \label{eq:cap_lloyd}
\end{equation}
If any part retains fewer than $10$ points during optimization, we discard the candidate decomposition for this $k$.
Because flat or bulbous single-part shapes are poorly approximated by a capsule alone, we evaluate a $k$-part partition using the weighted minimum-primitive residual
\begin{equation}
  e_{\text{fit}}^{(k)}=\sum_{j=1}^{k} w_j\min\!\big(e_j^{\text{cap}},\,e_j^{\text{box}},\,e_j^{\text{ell}}\big),
  \label{eq:prim_fit}
\end{equation}
where $w_j=|\mathcal{P}_j|/4{,}000$ denotes the point fraction of part $\mathcal{P}_j$, and $e_j^{\text{cap}}$, $e_j^{\text{box}}$, and $e_j^{\text{ell}}$ are the mean squared distances from the points in $\mathcal{P}_j$ to the surfaces of the fitted capsule, a PCA-aligned bounding box (clipped to the 2nd--98th percentiles along each principal axis), and a PCA ellipsoid sharing the same half-extents, respectively.
Starting from $k=1$, we greedily accept the $(k+1)$-part partition (up to three parts) if and only if three criteria are met: (i) substantial residual reduction ($e_{\text{fit}}^{(k+1)}\le 0.5\,e_{\text{fit}}^{(k)}$), (ii) non-trivial part size ($\min_j w_j\ge 0.10$), and (iii) non-collinear part axes ($|\mathbf{a}_i^{\top}\mathbf{a}_j|\le 0.9$ for all $i\ne j$); otherwise, splitting terminates.

\noindent\textbf{Canonical Axis Voting.}
Each part with at least $30$ points fits a PCA-aligned 2nd--98th percentile box with sorted widths $b_1\ge b_2\ge b_3$ along unit axes $(\mathbf{u}_1,\mathbf{u}_2,\mathbf{u}_3)$, casting an undirected major-axis vote along $\mathbf{u}_1$ with score $s_{\text{long}}=\frac{1}{5}w\log(b_1/b_2)$ and a normal vote along $\mathbf{u}_3$ with score $s_{\text{norm}}=\frac{1}{5}w\log(b_2/b_3)$ (when $k>1$, the whole object also casts votes weighted by $0.9$). Pooling proposals across all five seeds, unsigned directional clustering merges proposals within $10^\circ$ ($|\mathbf{u}_a^{\top}\mathbf{u}_b|\ge\cos 10^\circ$), summing their scores and taking the sign-aligned chordal mean direction:
\begin{itemize}
  \item \textbf{Primary axis $\mathbf{r}_1$.} If the strongest cluster reaches $s_1^\ast\ge\log 1.1\approx 0.095$, its mean direction defines $\mathbf{r}_1$; otherwise the object lacks a dominant axis and is marked \emph{isotropic} ($R=I_3$).
  \item \textbf{Secondary axis $\mathbf{r}_2$ and cross-section fallback.} Proposals not near-parallel to $\mathbf{r}_1$ ($|\mathbf{u}_i^{\top}\mathbf{r}_1|<0.5$) are projected onto the normal plane $\mathbf{r}_1^\perp$ as $\tilde{\mathbf{u}}_i=(I-\mathbf{r}_1\mathbf{r}_1^{\top})\mathbf{u}_i/\lVert(I-\mathbf{r}_1\mathbf{r}_1^{\top})\mathbf{u}_i\rVert$, reweighted by $\tilde{s}_i=s_i(1-(\mathbf{u}_i^{\top}\mathbf{r}_1)^2)$, and clustered within $10^\circ$. If the top cluster reaches $\tilde{s}_2^\ast\ge\log 1.1$, it defines $\mathbf{r}_2$. Otherwise, we project all points onto $\mathbf{r}_1^\perp$: if rotating the 2D cross-section by $45^\circ$ yields a 97th-percentile nearest-neighbor deviation below $0.06\,r_{95}$ ($r_{95}$ the 95th-percentile radius), the object is marked \emph{continuously symmetric} about $\mathbf{r}_1$; else $\mathbf{r}_2$ aligns with the minimum-area 2D bounding rectangle.
  \item \textbf{Right-handed completion.} The third axis is $\mathbf{r}_3=\mathbf{r}_1\times\mathbf{r}_2$, forming $R=[\mathbf{r}_1,\mathbf{r}_2,\mathbf{r}_3]\in SO(3)$.
\end{itemize}

\noindent\textbf{Axis Sign, Origin, and Bounding Extent.}
Along each axis $m\in\{1,2,3\}$, standardized projection skewness $\gamma_m$ fixes $+\mathbf{r}_m$ toward the longer tail whenever $|\gamma_m|\ge 0.25$ and leaves the sign unfixed otherwise. When majority voting across the five seeds indicates that $\mathbf{r}_1$ and $\mathbf{r}_2$ stem from distinct parts with centers $\mathbf{c}_1,\mathbf{c}_2$ (or when no seed votes and $\lVert\mathbf{c}_1-\mathbf{c}_2\rVert$ exceeds $5\%$ of object size), the canonical origin $\mathbf{o}$ is placed at the midpoint of the closest points between the two part axis lines $\{\mathbf{c}_1+\alpha\mathbf{r}_1\}$ and $\{\mathbf{c}_2+\beta\mathbf{r}_2\}$, naturally localizing structural junctions such as a mug or flask handle attachment. Otherwise, $\mathbf{o}$ is set to the primary-part center (or the global point mean when isotropic). Projecting all mesh vertices onto $R$ via $\mathbf{y}=R^{\top}(\mathbf{v}-\mathbf{o})$, the 2nd and 98th percentiles $[\ell_m^{\min},\ell_m^{\max}]$ along each coordinate define the trimmed box extent $e_m=\ell_m^{\max}-\ell_m^{\min}$ and canonical box center.

\noindent\textbf{Symmetry-Aware Frame Labels.}
Starting from $F_1=R$, we enumerate up to $16$ discrete equivalent orientations $F_m\in\mathcal{G}\subset SO(3)$ by combining: (i) runner-up primary clusters with $s\ge 0.8\,s_1^\ast$ and $|\mathbf{u}^{\top}\mathbf{r}_1|<0.9$, (ii) runner-up secondary clusters with $\tilde{s}\ge 0.8\,\tilde{s}_2^\ast$ separated by $>15^\circ$ from $\mathbf{r}_2$, (iii) $90^\circ$ cross-section turns when $\mathbf{r}_2$ comes from a minimum-area rectangle, and (iv) sign flips $\pm\mathbf{r}_m$ on axes with $|\gamma_m|<0.25$ (with $\mathbf{r}_3=\mathbf{r}_1\times\mathbf{r}_2$ preserving $\det F_m=+1$). Under continuous axial symmetry, orientations differing only by a rotation about $\mathbf{r}_1$ are merged, and $S_{\text{valid}}$ in \cref{eq:rotloss} is computed in closed form over axial turns about $\mathbf{r}_1$.

Most objects admit a compact label set: on our in-the-wild corpus~\cite{egodex}, $|\mathcal{G}|$ equals $1,2,4,$ and $8$ for $15,24,41,$ and $19\%$ of objects (under $1\%$ have other sizes; mean $|\mathcal{G}|=3.86$), $4.6\%$ are isotropic, $18.9\%$ are continuously symmetric, and $4.3\%$ place the origin at a part junction (with $0$--$11\%$ isotropic, $0$--$38\%$ continuously symmetric, mean $|\mathcal{G}|\in[2.55,4.30]$, and $0$--$11\%$ junction origins across studio sources).

\subsection{Model Design}
\noindent\textbf{Visual Backbone and Camera Embeddings.}
The trunk has eight transformer layers of width $384$ on top of the frozen DINOv3 ViT-L/16~\cite{simeoni2025dinov3} encoder, which processes $224\times224$ crops of the object and both hands alongside the resized full image. Class and patch tokens are projected from $1024$ to $256$ dimensions via precomputed PCA before entering the trunk. Each patch token receives the unit camera ray through its patch center and the class token receives the crop-center ray, concatenated with crop log-scale $\log(h/f_x)$ ($h$ half the crop side)~\cite{li2022cliff}. Six Fourier frequencies encode this 4D vector before a two-layer MLP maps it to width $384$ and adds it to the visual token.

\noindent\textbf{Hand and Optional Mesh Tokens.}
Each HaWoR~\cite{zhang2025hawor} joint token encodes 2D image coordinates, log depth, wrist-relative 3D offset, and local joint rotation, complemented by one global MANO parameter token per hand. When an optional reference mesh is supplied, $4{,}096$ randomly rotated surface points normalized by object diameter are encoded by PointNet++~\cite{qi2017pointnetplusplus} into $64$ tokens carrying a log-diameter embedding.

\noindent\textbf{Prediction Heads.}
The object head predicts an allocentric 6D rotation~\cite{Zhou_2019_CVPR} composed as $R=R_{\text{ray}}R_{\text{allo}}$ ($R_{\text{ray}}$ rotating the optical axis onto the crop-center ray), a 2D origin offset in crop units, log depth $s$ decoded as $z=0.1\,(f_x/h)\,e^s$, log extents, canonical box center, and two logits for \emph{isotropic} and \emph{continuous} symmetry. The hand head outputs zero-initialized residuals over global and $15$ joint rotations, MANO shape, 2D wrist shift, and multiplicative depth scale, while the SDF head regresses signed distances for all $799$ hand keypoints and vertex contact logits directly from each hand query.

\subsection{Training Objective}
The total training objective is a weighted sum (term weights in parentheses).

\noindent\textbf{Object Losses.}
The axis-set rotation loss of \cref{eq:rotloss} ($1.0$, $\mu=0.2$) is skipped on isotropic objects. The origin loss ($1.0$) sums the $L_1$ error of log depth and crop-normalized 2D pixel offset. The extent loss ($0.5$, after permuting target axes by $g^\ast\in\mathcal{P}_{24}$ from $S_{\text{any}}$) sums the $L_1$ log-extent error and half the $L_1$ canonical box-center error. Both continuous-symmetry flags use binary cross-entropy ($0.1$).

\noindent\textbf{Hand and Signed-Distance Losses.}
On studio samples with ground-truth MANO, supervision is evaluated in ground-truth canonical object coordinates (decoupling hand gradients from object-pose error): chordal rotation losses on global orientation ($0.5$) and $15$ joints ($0.5$), shape $L_1$ ($0.05$), 3D joint and vertex $L_1$ in decimeters ($1.0$ each), 2D joint reprojection ($0.2$), hand-presence BCE ($0.1$), plus two physical-plausibility terms ramped linearly from step $2{,}000$ to $4{,}000$---mean $|d|$ in cm over ground-truth contact vertices ($0.1$) and mean penetration depth in mm over all $778$ vertices ($0.1$). Across all sources, the SDF head minimizes mean absolute error in cm over the $799$ vertex and joint distances ($1.0$) and contact BCE against $|d|<5$\,mm ($1.0$). On in-the-wild annotations~\cite{egodex}, object and SDF terms are supervised in full while MANO parameter terms are omitted because ground-truth MANO mesh fits are unavailable.

\subsection{Implementation and Runtime}
We train for $40\text{k}$ steps on four RTX PRO~5000 GPUs ($\sim\!5.6$\,h, per-GPU batch size $256$) using AdamW (peak learning rate $8\times10^{-4}$, $1\text{k}$-step linear warmup, weight decay $0.05$, gradient clip $1.0$, dropout $0.1$, \texttt{bf16}). Each hand crop is dropped independently with probability $0.1$ during training; object crops are derived from 2D segmentation masks on in-the-wild training data and at test time, and from 2D ground-truth object projections on studio captures lacking precomputed masks. On a single RTX~4090, end-to-end inference takes $33$\,ms per image including live DINOv3 feature extraction ($3.1$\,ms, or $327$\,FPS, with cached DINOv3 features).

\section{Experimental Setup}
\label{sec:supp_eval}

\subsection{Benchmarks and Data Splits}
We evaluate 3D trajectory annotation and single-image abstraction prediction on 30\,FPS contact sub-benchmarks of egocentric HOT3D~\cite{banerjee2025hot3d} (held-out episodes) and third-person HO3D v3~\cite{hampali2020honnotate}.

\noindent\textbf{Annotation Protocol.}
Keyframes are sampled and annotated at 8\,FPS across the 18 HO3D v3 tracks with source videos and the 12 main HOT3D object tracks. Pipeline ablations are evaluated directly on unsmoothed 8\,FPS contact keyframes, whereas dense trajectory accuracy and smoothness are evaluated after 30\,FPS interpolation, palm-frame consensus, trajectory smoothing, and density filtering. Tracks yielding no accepted contact segments output zero dense frames and count against coverage.

\noindent\textbf{Abstraction-Model Protocol.}
The feed-forward predictor is trained on a balanced 2\,FPS egocentric set of roughly 0.4M samples combining S3AL in-the-wild annotations~\cite{egodex} (over half the data) with contact-filtered egocentric streams from public multi-camera studio captures~\cite{Zhan_2024_CVPR,liu2024taco,Liu_2022_CVPR,banerjee2025hot3d,rim2026show3d,Kwon_2021_ICCV,wang2025hocap}. It is evaluated across all contact frames of both 3D benchmarks: HOT3D~\cite{banerjee2025hot3d} tests in-domain generalization on strictly held-out episodes, whereas third-person HO3D v3~\cite{hampali2020honnotate} tests zero-shot transfer to unseen objects and external cameras (no third-person dataset, including HO3D v2/v3 and DexYCB~\cite{chao:cvpr2021}, is ever seen in training). On the held-out test split of the training pool, in-the-wild videos are split by manipulation task so test tasks are unseen, and studio captures are split by episode. For the qualitative EPIC-KITCHENS-100~\cite{damen2022rescaling} comparison (\cref{fig:epic_wild}), the \emph{Studio datasets} baseline uses the same architecture and 40k-step schedule with the in-the-wild sampling weight set to zero.

\subsection{Input Regimes}
We evaluate the annotation pipeline under two input regimes to separate optimization and verification performance from upstream perception noise. \emph{Oracle} inputs supply ground-truth MANO hands, benchmark object masks, ground-truth camera intrinsics, and DA3 relative depth scaled by ground-truth hands. \emph{Full-auto} inputs supply off-the-shelf HaWoR~\cite{zhang2025hawor} hands, SAM3~\cite{carion2025sam3segmentconcepts} masks propagated per track from a single category-prompted Gemini box/phrase call (shared as common input across all methods and accounting for $<$1\% of the \$0.0019/frame API cost on HO3D), and DA3 depth scaled by HaWoR hands, while retaining benchmark intrinsics and contact-segment windows for uniform scoring across baselines. All competing baselines receive identical RGB frames, masks, and intrinsics; FoundationPose and Any6D additionally receive DA3 depth calibrated to metric scale via ground-truth hands, whereas our single-image abstraction model uses no depth input.

\subsection{Evaluation Metrics}
\noindent\textbf{3D Object Accuracy and Coverage.}
Symmetric Chamfer distance (CD, cm) averages bidirectional nearest-neighbor distances over 10{,}000 uniformly sampled mesh surface points in metric camera coordinates; CD$^2$ (cm$^2$) sums the two directional mean squared distances to heavily penalize catastrophic outliers, following HOLD~\cite{fan2024hold} and gSDF~\cite{chen2023gsdf}. F@5 and F@10 report surface F-scores at 5 and 10\,mm, and $<$2 and $<$5 denote the percentage of evaluated frames with CD below 2 and 5\,cm, respectively. Standard pose metrics ADD and ADD-S~\cite{hinterstoisser2012model} (cm) are reported whenever the evaluated geometry is the ground-truth mesh. Each method is scored over the frames on which it produces outputs, accompanied by its coverage (Cov., \%). To evaluate our mesh-free abstraction directly without test-time mesh placement, we also report volumetric 3D oriented bounding box IoU and $L_2$ errors (cm) of the box center and canonical frame origin (\cref{sec:abs_def}), extracting canonical boxes and origins from baseline reconstructions via the same geometric rule.

\noindent\textbf{Temporal Smoothness.}
On continuous 30\,FPS output segments, we measure object angular-velocity error ($\omega$, $^\circ$/s) and translational acceleration error ($a$, m/s$^2$) against ground-truth trajectories, weighting segments by length and computing accelerations via second-order finite differences over frame timestamps.

\noindent\textbf{Hand Pose and Signed Distances.}
Hand accuracy is measured by camera-space MPJPE and MPVPE (mm), their Procrustes-aligned counterparts (PA-MPJPE, PA-MPVPE), and Obj.\ MPJPE---the joint error expressed in the predicted canonical object frame, which directly reflects hand-object relative pose consistency. Geometric contact precision, recall, and F1 compare posed hand-object proximity against ground-truth contact. For the signed-distance head, we report mean absolute error (MAE, mm) across all 799 hand keypoints (778 vertices plus 21 joints), Near MAE restricted to keypoints within 1\,cm of the ground-truth object surface, and contact F1 at the vertex ($|d|<5$\,mm) and whole-hand levels.

\subsection{Baselines and Mesh Placement}
FoundationPose~\cite{foundationposewen2024} tracks with either the GT mesh or our canonical SAM3D mesh (re-registering upon tracking loss on HO3D). Any6D~\cite{lee2025any6d}, ForeHOI~\cite{chen2026forehoifeedforward3dobject}, and HORT~\cite{chen2025hort} reconstruct their own object geometry; note that HORT is trained on HO3D and DexYCB, so its HO3D numbers are in-domain. DoAsIDo~\cite{paliwal2026doasido} requires ground-truth hands. MagicHOI~\cite{wang2025magichoi} initializes object cameras via structure-from-motion (\texttt{hloc} + COLMAP) at its native 6\,FPS rate before four-stage 3D object optimization and single-hand HaMeR~\cite{pavlakos2024reconstructing} alignment; tracks where SfM fails to reconstruct an object model yield no outputs. EasyHOI~\cite{liu2025easyhoi} optimizes an uncalibrated camera frame rather than metric benchmark coordinates, so we report its per-frame runtime in \cref{tab:ann_main} and $\mathrm{Sim}(3)$-aligned subset accuracy in \cref{sec:supp_results}. AgentSTAR~\cite{mazur2026agenticstar} is evaluated with Claude Opus 5.5 on a four-track 2\,FPS HOT3D subset due to its high per-frame API and compute cost (\cref{sec:supp_results}). For signed distances, we compare against an image-agnostic per-keypoint training prior (predicting each keypoint's fixed training-split statistics without visual input) and a single-frame reproduction of InterField~\cite{fan2023arctic} trained on our split with identical DINOv3 features and HaWoR hand inputs (best validation checkpoint).

When placing an external mesh (either our track's VLM-scaled SAM3D mesh or the GT mesh, both retaining their metric dimensions without rescaling by $\mathbf{e}$) into our predicted canonical frame at test time, discrete axis permutations between the mesh's native frame and our predicted frame are resolved automatically via 2D silhouette agreement with the object mask rather than ground-truth pose. Selecting this correspondence with ground-truth pose instead provides an oracle upper bound (\emph{GT axes} in \cref{tab:supp_model_full}), which changes symmetric surface alignment (ADD-S) by at most 0.1--0.2\,cm on both benchmarks and confirms that mask agreement reliably resolves axis correspondence.

\section{Additional Experimental Results}
\label{sec:supp_results}

\subsection{Comparison with SoTA Approaches}

\begin{figure*}[t]
\centering
\suppimg{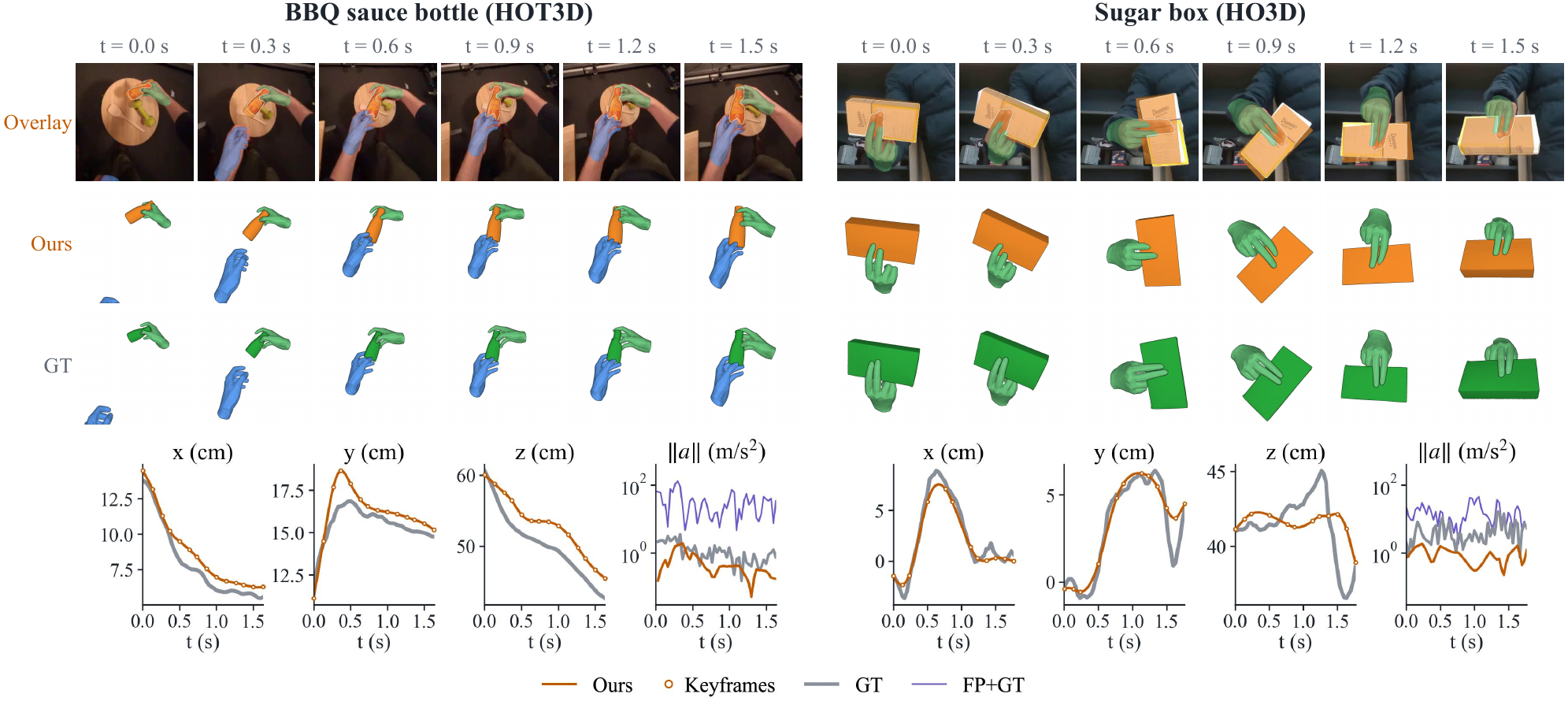}{\textwidth}
\caption{\textbf{Dense 30\,FPS trajectories} on HOT3D and HO3D v3 with full-auto inputs. Top: rendered frames of our smoothed output against ground truth. Bottom: object centroid trajectories (hollow circles mark accepted 8\,FPS keyframes) and acceleration profiles comparing our smoothed output against ground truth and FoundationPose with the GT mesh (\emph{FP+GT}).}
\label{fig:supp_temporal}
\vspace{-2mm}
\end{figure*}

\noindent\textbf{Dense 30 FPS Trajectories and Smoothness.}
\Cref{tab:temporal} evaluates 3D accuracy and temporal smoothness of the smoothed 30\,FPS trajectories derived from our 8\,FPS keyframes across HOT3D and HO3D v3 under both oracle and full-auto inputs, with representative sequences visualized in \cref{fig:supp_temporal}. Across all four settings, S3AL substantially outperforms every pose-tracking and monocular HOI baseline on the shared accepted frames---including in-domain HORT on HO3D and GT-hand DoAsIDo---showing that coupling verified canonical geometry with contact-guided refinement yields accurate metric trajectories even from noisy off-the-shelf detections. Beyond per-frame accuracy, palm-frame consensus markedly reduces angular-velocity error over \emph{w/o consensus} by suppressing discrete orientation flips on symmetric grasps, while trajectory smoothing cuts translational acceleration jitter by over $30\times$ on HOT3D and by an order of magnitude across frame-wise pose baselines.

\begin{table*}[t]
\centering
\caption{\textbf{Dense 30\,FPS trajectory accuracy and smoothness} on HOT3D and HO3D v3. Accuracy (CD in cm, CD$^2$ in cm$^2$, F@10) is evaluated on shared accepted contact frames; angular-velocity ($\omega$, $^\circ$/s) and acceleration ($a$, m/s$^2$) errors are evaluated over continuous 30\,FPS segments (baseline $\omega,a$ use the shared runs listed under Full-auto). $^\P$Uses GT hands. $^\S$Trained on HO3D and DexYCB. $^\ast$Point-cloud centroid only for $a$.}
\label{tab:temporal}
\small
\setlength{\tabcolsep}{4.5pt}
\begin{tabular*}{\textwidth}{@{\extracolsep{\fill}}l ccccc c ccccc@{}}
\toprule
& \multicolumn{5}{c}{Oracle inputs} & & \multicolumn{5}{c}{Full-auto inputs} \\
\cmidrule(lr){2-6}\cmidrule(l){8-12}
Method & CD$\downarrow$ & CD$^2\downarrow$ & F@10$\uparrow$ & $\omega\downarrow$ & $a\downarrow$ & & CD$\downarrow$ & CD$^2\downarrow$ & F@10$\uparrow$ & $\omega\downarrow$ & $a\downarrow$ \\
\midrule
\multicolumn{12}{@{}l}{\emph{HOT3D}} \\
FoundationPose (GT mesh)~\cite{foundationposewen2024} & 4.69 & 84.1 & .234 & -- & -- & & 5.06 & 102.1 & .220 & 616.0 & 42.90 \\
FoundationPose (SAM3D) & 5.08 & 97.7 & .215 & -- & -- & & 5.74 & 123.4 & .189 & 528.0 & 44.66 \\
Any6D~\cite{lee2025any6d} & 5.34 & 105.8 & .163 & -- & -- & & 6.08 & 171.1 & .152 & 973.5 & 60.01 \\
ForeHOI~\cite{chen2026forehoifeedforward3dobject} & 6.80 & 246.6 & .190 & -- & -- & & 5.76 & 129.6 & .189 & 1051.5 & 70.07 \\
DoAsIDo$^\P$~\cite{paliwal2026doasido} & 4.19 & 73.8 & .241 & -- & -- & & 3.97 & 66.6 & .253 & 87.1 & 38.39 \\
HORT$^{\S\ast}$~\cite{chen2025hort} & 3.88 & 74.0 & .238 & -- & -- & & 3.61 & 74.3 & .297 & -- & 46.72 \\
S3AL w/o consensus & -- & -- & -- & 133.5 & 1.16 & & -- & -- & -- & 111.1 & 1.26 \\
\textbf{S3AL} (dense) & \textbf{1.72} & \textbf{14.5} & \textbf{.500} & \textbf{72.9} & \textbf{1.15} & & \textbf{1.70} & \textbf{11.5} & \textbf{.470} & \textbf{63.8} & \textbf{1.23} \\
\midrule
\multicolumn{12}{@{}l}{\emph{HO3D v3}} \\
FoundationPose (GT mesh) & 3.32 & 69.4 & .343 & -- & -- & & 3.12 & 67.5 & .360 & 465.5 & 24.89 \\
FoundationPose (SAM3D) & 3.66 & 81.3 & .312 & -- & -- & & 3.59 & 86.1 & .327 & 521.8 & 23.80 \\
Any6D & 7.46 & 300.1 & .187 & -- & -- & & 8.06 & 541.0 & .189 & 674.0 & 56.78 \\
ForeHOI & 13.08 & 1431.2 & .161 & -- & -- & & 11.36 & 1072.5 & .164 & 832.7 & 59.51 \\
DoAsIDo$^\P$ & 2.55 & 32.3 & .335 & -- & -- & & 2.81 & 41.1 & .326 & 66.6 & 19.91 \\
HORT$^{\S\ast}$ & 3.82 & 43.3 & .166 & -- & -- & & 3.33 & 36.3 & .239 & -- & 7.27 \\
S3AL w/o consensus & -- & -- & -- & 79.2 & \textbf{1.65} & & -- & -- & -- & 79.4 & 1.68 \\
\textbf{S3AL} (dense) & \textbf{1.17} & \textbf{6.2} & \textbf{.612} & \textbf{55.5} & 1.68 & & \textbf{2.02} & \textbf{19.1} & \textbf{.456} & \textbf{52.2} & \textbf{1.64} \\
\bottomrule
\end{tabular*}
\vspace{-2mm}
\end{table*}

\noindent\textbf{Per-Frame Compute and API Cost Breakdown.}
\Cref{tab:ann_main} compares per-frame GPU runtime and API cost against competing pipelines on a single RTX~4090. Accounting for all VLM requests and GPU compute spent on rejected candidates, our full-auto pipeline costs \$0.0068 per retained 8\,FPS keyframe on HO3D v3 (\cref{tab:ablation_ann}), which amortizes after 30\,FPS interpolation and smoothing to \$0.0019 and 4.03\,GPU\,s per accepted frame (\cref{tab:ann_main})---using roughly an order of magnitude less GPU time per frame than per-sequence 3D fitting pipelines (38--138\,GPU\,s) and nearly $60\times$ less GPU time and over $300\times$ lower API cost than VLM-in-the-loop agentic tracking. Across our full-scale in-the-wild annotation run~\cite{egodex}, S3AL similarly averages \$0.0067 per accepted frame.

\noindent\textbf{Extended Analysis of MagicHOI, EasyHOI, and AgentSTAR.}
We evaluate three optimization-heavy baselines whose low SfM coverage, uncalibrated cameras, or prohibitive compute and API cost preclude full-benchmark accuracy comparison in \cref{tab:ann_main}.
First, MagicHOI~\cite{wang2025magichoi} requires object-centric structure-from-motion (\texttt{hloc} + COLMAP) before multi-stage 3D optimization. Under small foreground object crops (low effective object resolution on wide-angle egocentric views), heavy hand occlusion, and low-texture surfaces, COLMAP fails to reconstruct camera poses on 10 of 20 HO3D v3 tracks and 22 of 24 HOT3D tracks (\emph{Failed} in \cref{fig:ann_qual}), covering only 8.1\% ($487 / 5{,}983$) and 0.8\% ($47 / 6{,}160$) of contact frames at 73\,GPU\,s/frame on succeeded tracks. Even when SfM succeeds, it often yields underscaled or truncated meshes (\cref{fig:ann_qual}), averaging $3.39$\,cm CD on HO3D v3 and $10.44$\,cm on HOT3D, whereas full-auto S3AL reduces CD / CD$^2$ on the shared frames to $2.26$\,cm / $21.3$\,cm$^2$ and $1.70$\,cm / $10.4$\,cm$^2$, respectively.
Second, because EasyHOI~\cite{liu2025easyhoi} optimizes in an uncalibrated camera frame, aligning its native HOT3D predictions to ground-truth MANO hands via an oracle similarity transform ($\mathrm{Sim}(3)$) still leaves $16.15$\,cm overall CD, trailing full-auto S3AL on shared accepted frames ($14.87$ vs.\ $1.77$\,cm CD).
Third, on a four-track 2\,FPS HOT3D subset, AgentSTAR~\cite{mazur2026agenticstar} with Claude Opus 5.5 averages $9.22$\,cm CD across its outputs at 237\,GPU\,s and \$0.59 per frame, whereas non-VLM DoAsIDo and 2\,FPS S3AL reduce CD on the respective shared frames to $4.08$\,cm (vs.\ $8.13$\,cm) and $2.48$\,cm (vs.\ $5.62$\,cm).

\subsection{Extended Ablation Study of S3AL}
\label{sec:supp_vlm}
\begin{table*}[b]
\centering
\caption{\textbf{Extended 8\,FPS VLM verification ablations.} \textbf{(a)} Pipeline stage and contact-optimization ablation under oracle inputs on HO3D v3 keyframes (complementing full-auto \cref{tab:ablation_ann}; Cov.: keyframe retention rate \%, CD in cm, Cost: API cost per retained keyframe in USD, $^\ast$all outputs retained). \textbf{(b)} Replacing the default VLM auditor under full-auto inputs with identical zero-shot prompts (\emph{Tracks}: completed object tracks without API or schema failure; metrics averaged over completed tracks; CD in cm, CD$^2$ in cm$^2$).}
\label{tab:supp_vlm}
\small
\begin{minipage}[t]{0.487\textwidth}
\centering
\textbf{(a) Oracle-input ablation (HO3D v3)}\\[3pt]
\setlength{\tabcolsep}{3.8pt}
\begin{tabular*}{\linewidth}{@{\extracolsep{\fill}}l cccc@{}}
\toprule
Setting & Cov. & CD$\downarrow$ & CD$^2\downarrow$ & Cost (\$) \\
\midrule
\textbf{S3AL (full)} & 35.4 & \textbf{1.20} & \textbf{7.2} & 0.0072 \\
\midrule
w/o frame/pose review & 58.2 & 1.54 & 11.4 & 0.0026 \\
w/o final review & 41.8 & 1.52 & 14.7 & 0.0052 \\
w/o VLM$^\ast$ & 87.3 & 2.67 & 36.3 & 0 \\
\midrule
w/o contact optimization & 32.8 & 1.61 & 14.4 & 0.0078 \\
\bottomrule
\end{tabular*}
\end{minipage}\hfill
\begin{minipage}[t]{0.488\textwidth}
\centering
\textbf{(b) Choice of VLM auditor (full-auto inputs)}\\[3pt]
\setlength{\tabcolsep}{3.4pt}
\begin{tabular*}{\linewidth}{@{\extracolsep{\fill}}l ccccc@{}}
\toprule
Auditor VLM & Tracks & CD$\downarrow$ & CD$^2\downarrow$ & F@5$\uparrow$ & F@10$\uparrow$ \\
\midrule
\multicolumn{6}{@{}l}{\emph{HO3D v3}} \\
\texttt{gemini-3.8-flash} & \textbf{18/18} & \textbf{2.12} & \textbf{21.5} & \textbf{.254} & \textbf{.441} \\
\texttt{gpt-6-luna} & 15/18 & 2.33 & 25.8 & .230 & .410 \\
\texttt{qwen3.8-flash} & 13/18 & 2.87 & 31.9 & .173 & .315 \\
\midrule
\multicolumn{6}{@{}l}{\emph{HOT3D}} \\
\texttt{gemini-3.8-flash} & \textbf{12/12} & \textbf{1.93} & \textbf{16.2} & \textbf{.249} & \textbf{.441} \\
\texttt{gpt-6-luna} & 10/12 & 2.40 & 24.6 & .186 & .333 \\
\texttt{qwen3.8-flash} & 8/12 & 2.50 & 29.1 & .208 & .365 \\
\bottomrule
\end{tabular*}
\end{minipage}
\vspace{-2mm}
\end{table*}

\begin{table*}[t]
\centering
\caption{\textbf{Same-frame comparison without VLM verification} on smoothed 30\,FPS HO3D v3 trajectories (\$0 API cost). Accuracy (CD in cm, CD$^2$ in cm$^2$) is evaluated on the shared output frames of \emph{S3AL w/o VLM}. $^\dagger$Uses GT mesh and metric depth. $^\ddagger$Uses metric depth. $^\P$Uses GT hands. $^\S$Trained on HO3D and DexYCB.}
\label{tab:supp_novlm}
\small
\setlength{\tabcolsep}{4.8pt}
\begin{tabular*}{\textwidth}{@{\extracolsep{\fill}}l ccccc c ccccc@{}}
\toprule
& \multicolumn{5}{c}{Oracle inputs} & & \multicolumn{5}{c}{Full-auto inputs} \\
\cmidrule(lr){2-6}\cmidrule(l){8-12}
Method & CD$\downarrow$ & CD$^2\downarrow$ & F@5$\uparrow$ & F@10$\uparrow$ & $<$5$\uparrow$ & & CD$\downarrow$ & CD$^2\downarrow$ & F@5$\uparrow$ & F@10$\uparrow$ & $<$5$\uparrow$ \\
\midrule
FoundationPose (GT mesh)$^\dagger$~\cite{foundationposewen2024} & 4.68 & 125.5 & .145 & .268 & 74.7 & & 4.60 & 123.7 & .149 & .275 & 75.3 \\
FoundationPose (SAM3D)$^\ddagger$~\cite{foundationposewen2024} & 5.30 & 153.2 & .131 & .239 & 71.5 & & 5.21 & 150.8 & .135 & .246 & 72.8 \\
Any6D$^\ddagger$~\cite{lee2025any6d} & 8.90 & 474.2 & .079 & .151 & 46.9 & & 8.80 & 478.0 & .081 & .154 & 47.8 \\
ForeHOI~\cite{chen2026forehoifeedforward3dobject} & 16.62 & 2154.9 & .060 & .115 & 36.6 & & 16.99 & 2251.1 & .061 & .117 & 36.5 \\
DoAsIDo$^\P$~\cite{paliwal2026doasido} & 3.79 & 70.1 & .138 & .267 & 80.8 & & 3.70 & 67.1 & .140 & .271 & 82.2 \\
HORT$^\S$~\cite{chen2025hort} & 3.26 & 36.4 & .138 & .254 & 79.7 & & 3.24 & 36.3 & .141 & .259 & 80.0 \\
\midrule
\textbf{S3AL w/o VLM} & \textbf{2.66} & \textbf{34.8} & \textbf{.207} & \textbf{.369} & \textbf{88.3} & & \textbf{2.79} & \textbf{33.6} & \textbf{.193} & \textbf{.346} & \textbf{87.5} \\
\bottomrule
\end{tabular*}
\vspace{-2mm}
\end{table*}

\begin{table*}[!b]
\centering
\caption{\textbf{Extended single-image 3D object evaluation} on HOT3D (in-domain) and HO3D v3 (zero-shot), complementing \cref{tab:model}. \textbf{(a)} Direct evaluation of predicted 3D oriented bounding boxes and canonical origins without test-time mesh placement for our model (Cov.: output coverage \%; IoU: volumetric 3D box IoU; Center / Origin: $L_2$ error in cm; baseline boxes and origins extracted via \cref{sec:abs_def}). \textbf{(b)} Tight surface recall ($<$2\,cm in \%), classical ADD / ADD-S (cm), and oracle axis-selection upper bound (\emph{GT axes}) with GT meshes. $^\dagger$Uses GT mesh and metric depth. $^\ddagger$Uses metric depth. $^\P$Uses GT hands. $^\S$Trained on HO3D and DexYCB.}
\label{tab:supp_model_full}
\small
\setlength{\tabcolsep}{4.6pt}
\begin{tabular*}{\textwidth}{@{\extracolsep{\fill}}l cccc c cccc@{}}
\toprule
& \multicolumn{4}{c}{HOT3D (in-domain)} & & \multicolumn{4}{c}{HO3D v3 (zero-shot)} \\
\cmidrule(lr){2-5}\cmidrule(l){7-10}
\multicolumn{10}{@{}l}{\textbf{(a) Direct 3D oriented box and canonical origin evaluation} (no test-time mesh placement for our model)} \\[1pt]
Method & Cov. & IoU$\uparrow$ & Center$\downarrow$ & Origin$\downarrow$ & & Cov. & IoU$\uparrow$ & Center$\downarrow$ & Origin$\downarrow$ \\
\midrule
FoundationPose (GT mesh)$^\dagger$~\cite{foundationposewen2024} & 100 & .163 & 9.26 & 9.19 & & 100 & .194 & 7.69 & 7.82 \\
FoundationPose (SAM3D)$^\ddagger$~\cite{foundationposewen2024} & 90.5 & .137 & 8.98 & 9.20 & & 74.9 & .182 & 7.99 & 8.17 \\
Any6D$^\ddagger$~\cite{lee2025any6d} & 100 & .104 & 11.2 & 11.3 & & 100 & .088 & 13.4 & 13.1 \\
ForeHOI~\cite{chen2026forehoifeedforward3dobject} & 100 & .102 & 13.8 & 14.3 & & 100 & .076 & 21.4 & 21.1 \\
DoAsIDo$^\P$~\cite{paliwal2026doasido} & 54.3 & .151 & 7.93 & 7.73 & & 100 & .229 & 5.80 & 5.75 \\
HORT$^\S$~\cite{chen2025hort} & 99.4 & .094 & 8.37 & 8.42 & & 100 & .174 & 6.15 & 6.30 \\
\textbf{Ours} & 100 & .358 & \textbf{3.14} & \textbf{2.84} & & 100 & .149 & 6.94 & 7.09 \\
\textbf{Ours + mesh input} & 100 & \textbf{.372} & 3.26 & 3.00 & & 100 & \textbf{.279} & \textbf{4.03} & \textbf{4.20} \\
\midrule
\multicolumn{10}{@{}l}{\textbf{(b) Tight surface recall, classical pose metrics, and symmetry upper bound} (with GT meshes)} \\[1pt]
Method & F@5$\uparrow$ & $<$2$\uparrow$ & ADD$\downarrow$ & ADD-S$\downarrow$ & & F@5$\uparrow$ & $<$2$\uparrow$ & ADD$\downarrow$ & ADD-S$\downarrow$ \\
\midrule
FoundationPose$^\dagger$~\cite{foundationposewen2024} & .126 & 28.8 & 12.30 & 6.58 & & .134 & 33.8 & 9.70 & 5.15 \\
\textbf{Ours} (silhouette) & .308 & 74.5 & 7.38 & 1.75 & & .117 & 22.2 & 10.69 & 4.35 \\
\textbf{Ours + mesh input} (silhouette) & \textbf{.322} & 75.9 & 7.30 & 1.87 & & \textbf{.180} & \textbf{46.9} & 8.27 & \textbf{2.59} \\
\cmidrule(lr){1-10}
\emph{Ours (GT axes)} & .322 & \textbf{77.1} & \textbf{3.18} & \textbf{1.56} & & .114 & 24.9 & \textbf{7.55} & 4.25 \\
\bottomrule
\end{tabular*}
\vspace{-2mm}
\end{table*}

\noindent\textbf{Oracle-Input Stage and Contact Ablation.}
\Cref{tab:supp_vlm}(a) complements the full-auto ablation in \cref{tab:ablation_ann} of the main paper by evaluating the same variants under oracle inputs (ground-truth hands, masks, and hand-calibrated depth scale) on unsmoothed 8\,FPS HO3D v3 keyframes, isolating verification and pose optimization from upstream perception noise. Even with clean inputs, dropping all VLM stages (\emph{w/o VLM}) inflates outlier-sensitive CD$^2$ fivefold as broken meshes and symmetric rotation flips go unfiltered, and skipping only final review (\emph{w/o final review}) still more than doubles CD$^2$. Replacing upstream frame and pose reviews with heuristics (\emph{w/o frame/pose review}) lowers raw API cost, but feeds poor rotation candidates into optimization that final review only partially catches. Conversely, removing contact-guided refinement (\emph{w/o contact optimization}) doubles CD$^2$ while lowering final-review acceptance and raising effective API cost per retained keyframe, confirming that continuous contact optimization is needed even with oracle hands and masks to align discrete rotation hypotheses into valid, verifiable contact.

\noindent\textbf{Choice of VLM Auditor.}
\Cref{tab:supp_vlm}(b) evaluates our full-auto pipeline on unsmoothed 8\,FPS keyframes across HO3D v3 and HOT3D when replacing our default auditor (\texttt{gemini-3.8-flash}) with \texttt{gpt-6-luna} or \texttt{qwen3.8-flash} under identical zero-shot prompt templates (\cref{sec:supp_prompts}). Without per-model prompt tuning, all three auditors improve 3D accuracy over \emph{w/o VLM} (\cref{tab:ablation_ann}) and surpass external baselines, confirming that our fixed-stage verification protocol is not tied to a single proprietary model. Across both benchmarks, \texttt{gemini-3.8-flash} achieves the highest accuracy on every metric while completing all object tracks reliably, whereas \texttt{gpt-6-luna} and \texttt{qwen3.8-flash} drop several tracks due to persistent API errors or malformed structured-schema responses. Restricting evaluation strictly to the shared completed tracks confirms that \texttt{gemini-3.8-flash} also provides stronger visual-spatial discrimination head-to-head (\eg, $2.07$ vs.\ $2.33$\,cm CD against \texttt{gpt-6-luna} on HO3D v3, and $1.81$ vs.\ $2.50$\,cm against \texttt{qwen3.8-flash} on HOT3D, where the latter is overly permissive during final review).

\noindent\textbf{Same-Frame Comparison Without VLM Verification.}
To verify that S3AL's accuracy gains stem from geometric and contact-guided pose optimization rather than VLM cherry-picking, \cref{tab:supp_novlm} compares \emph{S3AL w/o VLM} (\$0 API cost) against all baselines on its unscreened 30\,FPS HO3D v3 output frames under both oracle and full-auto inputs. Across both input regimes, our unverified pipeline outperforms every baseline on all five accuracy metrics---including depth-assisted FoundationPose with the ground-truth mesh, GT-hand DoAsIDo, and in-domain HORT. Its smoothed full-auto trajectories also preserve strong temporal coherence without VLM filtering ($67.2^\circ\text{/s}$ angular-velocity and $1.52\,\text{m/s}^2$ acceleration error on HO3D v3; $79.7^\circ\text{/s}$ and $2.20\,\text{m/s}^2$ on HOT3D), maintaining an order-of-magnitude lower acceleration jitter than frame-wise baselines (\cref{tab:temporal}). Fixed-stage VLM verification (\cref{tab:ablation_ann,tab:supp_vlm}) then removes the remaining tail failures for high-purity dataset curation.

\subsection{Extended Analysis of Mesh-Free Abstraction}
\begin{table*}[!b]
\centering
\caption{\textbf{Hand refinement and signed-distance evaluation} on the held-out test split. \textbf{(a)} Camera-space and Procrustes-aligned joint/vertex errors (mm, \emph{top}), alongside keypoint signed-distance MAE (mm; \emph{Near}: within 1\,cm of the object surface) and contact F1 at $|d|<5$\,mm (\emph{bottom}; $^\dagger$vertices only, $^\ast$thresholded distances). \textbf{(b)} Disaggregated vertex signed-distance MAE (mm over the 778 MANO vertices) of the same jointly trained model across the two subsets of the held-out test split---unseen in-the-wild tasks~\cite{egodex} and held-out studio episodes---against an image-agnostic per-keypoint training prior and our single-frame reproduction of InterField~\cite{fan2023arctic}.}
\label{tab:hands}
\footnotesize
\begin{minipage}[t]{0.82\textwidth}
\centering
\textbf{(a) Overall hand refinement \& signed-distance accuracy}\\[3pt]
\setlength{\tabcolsep}{2.1pt}
\begin{tabular*}{\linewidth}{@{\extracolsep{\fill}}lcccc@{}}
\toprule
Hands & MPJPE$\downarrow$ & MPVPE$\downarrow$ & PA-MPJPE$\downarrow$ & PA-MPVPE$\downarrow$ \\
\midrule
HaWoR~\cite{zhang2025hawor} (input) & 71.3 & 70.9 & 8.9 & 8.6 \\
\textbf{Ours} & 26.6 & 26.3 & \textbf{5.8} & \textbf{5.5} \\
\textbf{Ours + mesh input} & \textbf{26.4} & \textbf{26.1} & 5.9 & 5.6 \\
\midrule
Signed distances & MAE$\downarrow$ & Near MAE$\downarrow$ & Vertex F1$\uparrow$ & Hand F1$\uparrow$ \\
\midrule
Training prior & 11.21 & 25.64 & .283 & .755 \\
InterField~\cite{fan2023arctic} (repr.) & 5.49$^\dagger$ & 9.52 & .356$^\ast$ & -- \\
\textbf{Ours} & 4.02 & 6.48 & .510 & .925 \\
\textbf{Ours + mesh input} & \textbf{3.78} & \textbf{6.13} & \textbf{.524} & \textbf{.930} \\
\bottomrule
\end{tabular*}
\end{minipage}

\vspace{6pt}
\begin{minipage}[t]{0.82\textwidth}
\centering
\textbf{(b) Vertex signed-distance MAE by test subset (mm)}\\[3pt]
\setlength{\tabcolsep}{2.8pt}
\begin{tabular*}{\linewidth}{@{\extracolsep{\fill}}lrccc@{}}
\toprule
Test subset & Hands & Prior & InterField & \textbf{Ours} \\
\midrule
In-the-wild tasks~\cite{egodex} & 39{,}988 & 12.07 & 5.92 & \textbf{4.54} \\
Studio episodes & 55{,}987 & 10.59 & 5.18 & \textbf{3.64} \\
\midrule
All test & 95{,}975 & 11.20 & 5.49 & \textbf{4.01} \\
\bottomrule
\end{tabular*}
\end{minipage}
\vspace{-2mm}
\end{table*}

\noindent\textbf{Direct Mesh-Free 3D Box and Origin Evaluation.}
\Cref{tab:supp_model_full}(a) evaluates our object abstraction directly in its native parameterization---3D oriented bounding box IoU, box-center error, and canonical-origin error---without placing external meshes or searching over 2D mask silhouettes at test time. On in-domain HOT3D, our mesh-free predictor more than doubles 3D box IoU over depth-assisted FoundationPose with ground-truth meshes ($.358$ vs.\ $.163$, and nearly $4\times$ HORT's $.094$) while cutting center and origin errors by over two-thirds ($3.14$ / $2.84$ vs.\ $9.26$ / $9.19$\,cm). On zero-shot third-person HO3D v3, unconditioned \emph{Ours} still achieves lower center and origin errors than depth-assisted FoundationPose ($6.94$ / $7.09$ vs.\ $7.69$ / $7.82$\,cm) despite the egocentric-to-third-person depth shift, and supplying optional reference-mesh tokens (\emph{Ours + mesh input}) yields the highest 3D box IoU ($.279$) and lowest localization error ($4.03$ / $4.20$\,cm) across all baselines, outperforming both in-domain HORT and GT-hand DoAsIDo.

\noindent\textbf{GT-Mesh Pose Metrics and Symmetry Analysis.}
\Cref{tab:supp_model_full}(b) complements \cref{tab:model} of the main paper by reporting tight surface recall (F@5 and $<$2\,cm), classical pose metrics (ADD and ADD-S), and the \emph{GT axes} oracle correspondence upper bound with ground-truth meshes. On in-domain HOT3D, our single-image model more than doubles strict $<$2\,cm recall and cuts ADD-S to nearly one-fourth of depth-assisted FoundationPose. Comparing \emph{Ours (silhouette)} against \emph{Ours (GT axes)} further isolates the effect of object symmetry: oracle axis selection substantially reduces asymmetric ADD while leaving symmetric ADD-S nearly unchanged, confirming that residual ADD error stems from visually indistinguishable discrete symmetries under grasp occlusion rather than geometric misplacement. On zero-shot third-person HO3D v3, supplying optional reference-mesh tokens (\emph{Ours + mesh input}) resolves cross-view scale-depth ambiguity and achieves the strongest surface recall and ADD-S.

\noindent\textbf{Hand Refinement and Signed-Distance Accuracy.}
\Cref{tab:hands}(a) evaluates hand refinement and dense keypoint signed-distance prediction on the held-out test split. Conditioning on joint object-hand visual tokens cuts camera-space hand error by 63\% over input HaWoR estimates while also improving Procrustes-aligned articulation, showing that object context anchors monocular hand depth. For signed distances, conditioning jointly on target-object and hand tokens enables our head to substantially outperform both the empirical training prior and our in-domain DINOv3 reproduction of InterField~\cite{fan2023arctic} across overall MAE, near-surface MAE (within 1\,cm of the object), and contact F1, with optional mesh tokens (\emph{Ours + mesh input}) providing a further gain. \Cref{tab:hands}(b) confirms that this gain holds across both unseen in-the-wild tasks and studio episodes.

\noindent\textbf{In-the-Wild vs.\ Studio Subsets.}
Disaggregating our jointly trained model across the two subsets of the held-out test split ($20{,}746$ frames from unseen in-the-wild manipulation tasks~\cite{egodex} and $29{,}718$ frames from held-out multi-camera studio episodes), object placement accuracy on unseen in-the-wild tasks closely matches studio episodes ($1.14$ vs.\ $1.22$\,cm CD, $.628$ vs.\ $.633$ F@10, and $85.7\%$ vs.\ $85.5\%$ $<$2\,cm recall), confirming that S3AL annotations provide geometrically coherent 3D supervision on par with studio ground truth.

\FloatBarrier
\section{Limitations and Future Work}
\label{sec:supp_limits}

Our current framework focuses on rigid objects, leaving articulated (\eg, scissors, laptops) and deformable (\eg, cloth, sponges) items outside our present scope; extending our part-based canonical abstraction to articulated kinematic chains is a natural direction. In addition, under rapid in-hand reorientation or prolonged occlusion, stronger temporal tracking and contact priors could further improve continuous pose recovery, while explicit focal-length and viewpoint conditioning could strengthen cross-camera transfer.
\section{VLM Verification Prompts}
\label{sec:supp_prompts}

\newenvironment{promptblock}{\par\vspace{1pt}\begingroup\ttfamily\scriptsize\raggedright\setlength{\parindent}{0pt}%
  \setlength{\parskip}{0pt}}{\endgroup\par\vspace{1pt}}
\newcommand{\promptline}[1]{\hangindent=1.5em\hangafter=1 #1\par}
\newcommand{\promptgap}{\vspace{0.35\baselineskip}}

All datasets and benchmarks share the same unified open-vocabulary prompt templates, executed with \texttt{gemini-3.8-flash} using schema-enforced JSON outputs at temperature $0$ (low thinking for frame, mesh, and scale review; medium thinking for pose and final review). Placeholders in braces supply only the detected object label, frame identifiers, and calibrated depth summary at runtime. Below we summarize the core prompt instructions for each of the four fixed verification stages, omitting JSON formatting boilerplate and schema field definitions.

\subsection{Frame Review}
Evaluates batches of candidate frames with red mask outlines (fixing physical object identity to \texttt{\{reference\_\allowbreak frame\}} after the first batch, and re-ranking anchor candidates when exceeding the mesh budget).
\begin{promptblock}
\promptline{Select reconstruction anchors and screen per-frame masks for one rigid physical object: \{object\_\allowbreak{}name\} (Frames: \{frame\_\allowbreak{}ids\}). Red outlines are candidate masks, not trusted labels.}
\promptline{1. Geometric checks (evaluated independently of contact): mark same\_\allowbreak{}physical\_\allowbreak{}object false if the track switches to a different physical instance; mark mask\_\allowbreak{}usable true when the visible target pixels are cleanly segmented for pose estimation (normal hand occlusion is allowed; reject wrong targets, merged hand/background, or severe leakage); mark can\_\allowbreak{}generate\_\allowbreak{}mesh true only when visible geometry is sufficiently complete to reconstruct a 3D mesh.}
\promptline{2. Interaction check: from source RGB rather than 2D overlap alone, classify target\_\allowbreak{}interaction as contact (at least one hand holds, supports, presses, or touches this target), no\_\allowbreak{}contact, other\_\allowbreak{}object, or uncertain, and identify hand\_\allowbreak{}association in \{left, right, both, neither, unknown\}. Require explicit contact in each frame individually.}
\promptline{3. Anchor ranking: rank all frames passing the three geometric checks by suitability for 3D mesh generation, preferring clear complete views with minimal hand occlusion.}
\end{promptblock}

\subsection{Mesh and Scale Review}
\noindent\textbf{Mesh Review.} Evaluates a two-row grid repeating the anchor RGB/mask alongside six multi-view candidate SAM3D renders.
\begin{promptblock}
\promptline{Review the generated mesh of \{object\_\allowbreak{}name\} across six rendered views covering opposite faces, sides, and oblique directions (PCA axis signs are arbitrary; ignore texture or color differences). Inspect all six views before judging hidden features. Reject meshes with missing functional parts, merged hand or background geometry, debris, or wrong global shape, and accept \{anchor\_\allowbreak{}frame\} if the 3D geometry is usable.}
\end{promptblock}

\noindent\textbf{Scale Estimation.} Estimates metric dimensions from five panels (full RGB scene, target crop, hand-calibrated metric depth map, orthographic canonical mesh axes \texttt{A/B/C}, and mask-fitted axis projection) and calibrated depth statistics.
\begin{promptblock}
\promptline{Estimate the complete physical dimensions (in meters) of \{object\_\allowbreak{}name\} using the RGB context, familiar hand/scene size references, calibrated depth cues, and canonical mesh axes A=X, B=Y, C=Z:}
\promptline{1. Reconcile partial visible depth with category priors to estimate the complete object including occluded or frame-truncated parts.}
\promptline{2. Return longest\_\allowbreak{}dimension\_\allowbreak{}m and the full metric extent along each canonical mesh axis A, B, C independently. Mark an axis supported true when its correspondence and metric extent are visually grounded (including thin structures such as plates, cards, or lids whose mesh thickness is inflated), and false otherwise so raw mesh proportions are preserved.}
\end{promptblock}

\subsection{Pose Review}
Evaluates candidate rotations in camera-view triptychs (\texttt{LEFT}: observed RGB, \texttt{MIDDLE}: rendered mesh overlay, \texttt{RIGHT}: mesh only).
\begin{promptblock}
\promptline{Assess whether the candidate 3D orientation R of \{object\_\allowbreak{}name\} agrees with the observed RGB across the triptych panels:}
\promptline{1. Compare (i) principal structural axis direction and surface-plane foreshortening, (ii) visible asymmetric cues (open/closed end, handle, functional surface, or tip), and (iii) part arrangement relative to the interacting hand.}
\promptline{2. Keep rotation, translation, scale, and mesh-shape defects separate: do not mistake filled cavities, missing thin features, or initial translation offsets for wrong R, and allow symmetric spin when all observable cues agree.}
\promptline{3. Classify orientation\_\allowbreak{}adequacy as adequate (observable structural directions and asymmetric cues agree up to minor tilt), inadequate (clear rotational contradiction between source RGB and rendered mesh), or unobservable, and return decision in \{accept, reject, uncertain\}, accepting only adequate orientation with usable geometry.}
\end{promptblock}

\subsection{Final Review}
Evaluates optimized poses using the camera-view triptych together with orthogonal \texttt{contact\_side} and \texttt{contact\_top} renders of the 3D hand--object pair.
\begin{promptblock}
\promptline{Assess whether the reconstructed 3D hand-object pair preserves the core interaction visible in source RGB, using the camera triptych for hand ownership and alignment and the orthogonal contact\_\allowbreak{}side / contact\_\allowbreak{}top views to check for detached grasps, spurious contact, or gross interpenetration hidden in the camera projection:}
\promptline{- Tier A (accept): observed interaction, object orientation, scale, and alignment agree cleanly with no notable defect.}
\promptline{- Tier B (accept): core grasp/support and observable orientation are preserved, with only minor local shape, scale, silhouette, or individual-finger contact residuals.}
\promptline{- Tier C (reject): lost grasp/support, wrong hand association, spurious overlap without true grasp, or major orientation/shape/scale error.}
\promptline{- Tier U (reject): visual evidence cannot establish the core hand-object relation or rule out a major defect.}
\end{promptblock}

\end{document}